\documentclass{article}

\usepackage{arxiv}

\usepackage{cite}
\usepackage{amsmath,amssymb,amsfonts}
\usepackage{algorithmic}
\usepackage{graphicx}
\usepackage{textcomp}
\usepackage{xcolor}
\usepackage[percent]{overpic}
\usepackage{paralist}
\usepackage{booktabs}
\usepackage{colortbl}
\usepackage{csquotes}
\usepackage[bookmarks=false]{hyperref}
\usepackage{fancyhdr}

\usepackage{amsthm}
\theoremstyle{definition}
\newtheorem{defn}{Definition}%[section]

\usepackage{hyperref}
\usepackage{subcaption}

\begin{document}

\title{Empirical Study on\\ the Benefits of Multiobjectivization for\\ Solving Single-Objective Problems}

\author{
  Vera Steinhoff\\
  Statistics and Optimization\\
  University of M{\"u}nster \\
  M{\"u}nster, Germany \\
  \texttt{v.steinhoff@uni-muenster.de} \\
  \And
  Pascal Kerschke \\
  Statistics and Optimization\\
  University of M{\"u}nster \\
  M{\"u}nster, Germany \\
  \texttt{kerschke@uni-muenster.de}
   \And
  Christian Grimme \\
  Statistics and Optimization\\
  University of M{\"u}nster \\
  M{\"u}nster, Germany \\
  \texttt{christian.grimme@uni-muenster.de}
}

\maketitle

%\thispagestyle{fancy}
%\lfoot{\vspace*{-1.25cm}\rule{\columnwidth}{0.2pt}\\\footnotesize \textcopyright2020 IEEE. Personal use of this material is permitted. Permission from IEEE must be obtained for all other uses, in any current or future media, including reprinting/republishing this material for advertising or promotional purposes, creating new collective works, for resale or redistribution to servers or lists, or reuse of any copyrighted component of this work in other works.\\ This version has been accepted for publication at the \textit{International Joint Conference on Neural Networks (IJCNN)} 2020, which is part of the \textit{IEEE World Congress on Computational Intelligence (IEEE WCCI)} 2020.}\cfoot{}

\begin{abstract}
When dealing with continuous single-objective problems, multimodality poses one of the biggest difficulties for global optimization. 
Local optima are often preventing algorithms from making progress and thus pose a severe threat.
In this paper we analyze how single-objective optimization can benefit from multiobjectivization by considering an additional objective.
With the use of a sophisticated visualization technique based on the multi-objective gradients, the properties of the arising multi-objective landscapes are illustrated and examined. 
We will empirically show that the multi-objective optimizer MOGSA is able to exploit these properties to overcome local traps.
The performance of MOGSA is assessed on a testbed of several functions provided by the COCO platform. The results are compared to the local optimizer Nelder-Mead. 
\end{abstract}


\section{Introduction}
%Whether it is in the industry where the production costs should be minimized or in the economy during a decision for a car taking into account the speed, the price, and the emissions, everywhere the global optimal solution or the set of global optimal solutions is sought. 
% numerous experiments in black-box setting were published to demonstrate suitability of these search methods/procedures in multimodal landscapes
Optimization is essentially everywhere.
Whether it is in the field of production, logistics, in medicine or biology; everywhere the global optimal solution or the set of global optimal solutions is sought.
However, most real-world problems are of non-linear nature and naturally multimodal which poses severe problems to global optimization. 
Multimodality, the existence of multiple (local) optima, is regarded as one of the biggest challenges for continuous single-objective problems~\cite{preuss2015}.
A lot of algorithms get stuck searching for the global optimum or are requiring many function evaluations to escape local optima.
One of the most popular strategies for dealing with multimodal problems are population-based methods like evolutionary algorithms due to their global search abilities~\cite{beyer2001ES}.

In this paper we will examine another approach of coping with local traps, namely \emph{multiobjectivization}.
By transforming a single-objective into a multi-objective problem, we aim at exploiting the properties of multi-objective landscapes. % for finding the global single-objective optimum.
So far, the characteristics of single-objective optimization problems have often been directly transferred to the multi-objective domain. 
Due to the existence of multiple objectives to be optimized, visualization seemed to be much more difficult.
Thus, multimodality was regarded as an even bigger challenge for multi-objective problems.
However, as empirically shown and hypothesized in papers by Kerschke and Grimme~\cite{GrimmeKT2019Multimodality}, local optima in continuous multi-objective optimization should not be regarded as challenges but as chances for finding the global efficient set.
By visualizing multi-objective problems in gradient-based heatmaps, insights into the landscape and specific characteristics such as basins of attraction and efficient sets could be gained~\cite{MOlandscapes,kerschke2018search}. 
% understanding
%Visualizing multi-objective problems in gradient-based heatmaps provide insights into the landscape and specific characteristics such as basins of attraction and efficient sets~\cite{MOlandscapes}. 
Furthermore, local efficient sets were observed to be cut by ridges whereas global efficient sets should be ridge-free.
Based on these observations, the Multi-Objective Gradient Sliding Algorithm (MOGSA) was developed to follow the path of the multi-objective gradients and explore the efficient sets to finally find the globally efficient one~\cite{GrimmeKT2019Multimodality}.
Instead of getting trapped in a local optima, MOGSA crosses ridges and thereby efficiently steers into more promising basins of attraction.
%slides from basin to basin in
%they can guide the search towards the global efficient set.
%To overcome the problem that goes along with multimodality in the single-objective case, our work is based on previous research of Kerschke and Grimme~\cite{GrimmeKT2019Multimodality,MOlandscapes}.
To overcome the problems caused by multimodality in the single-objective case, we build on previous research of Kerschke and Grimme~\cite{GrimmeKT2019Multimodality,MOlandscapes} and use a multi-objective optimizer to tackle single-objective problems.
\begin{center}
\vspace*{-0.5cm}
\hspace*{-0.1cm}\begin{tabular}{|p{16.2cm}|}
    \hline
      \rule{0pt}{13pt}\emph{Contribution:} The local optimizer MOGSA will be applied to a set of initially single-objective problems, which were transformed into multi-objective ones by adding an additional objective. We show how locality in the multi-objective landscape can guide the algorithm towards the global optimum of the single-objective problem. Our aim is not to challenge existing algorithms but to show that multi-objective methods can help in overcoming the difficulties of multimodal single-objective optimization. Our experiments, which are based on the well-known black-box optimization benchmark (BBOB)~\cite{Definition_noislessFunctions} -- a suite of continuous test functions with various difficulties -- reveal that MOGSA is able to solve even highly multimodal single-objective problems. \\[-0.75em]
\rule{0pt}{1pt}\\
    \hline
    \end{tabular}
\end{center}

%can even be exploited hope to reduce local traps and find the global single-objective optimum by exploiting the multi-objective landscape.
For our experiments, we make use of the COCO (COmparing Continuous Optimizers) platform~\cite{hansen2016cocoplat}, which provides tools for performance assessment, as well as access to the aforementioned BBOB test suite.
The performance of the multi-objective local optimizer MOGSA is compared to the single-objective local optimizer Nelder-Mead~\cite{nelder1965simplex}.
%To compare the results generated for different runs of the local multi-objective optimizer MOGSA on these functions, the local single-objective optimizer Nelder-Mead~\cite{nelder1965simplex} is run on the same problems. 
% challenge existing approaches% belonging to the standard repertoire of optimization practitioners; heuristic nature S.10
To further understand the effect of multiobjectivization, we will visually analyze the corresponding landscapes and study MOGSA's search behavior on the problems.
% analyze the effect of multiobjectivization for multimodal single-objective problems.

In Section~\ref{sec:background} we will formally define single-objective and multi-objective problems, as well as their properties. Furthermore, we will give an overview of related work on multiobjectivization. Afterwards, our concept is explained in detail in Section~\ref{sec:concept}. 
Subsection~\ref{sec:SOtoMO} describes and explains our procedure of transforming single-objective into multi-objective problems. Subsection~\ref{sec:MOlandscape} underlines this approach by means of a visual investigation of the resulting multi-objective landscape. 
The local optimizer MOGSA and our adaptations to it are described conceptually in the following Subsection~\ref{sec:mogsa}. 
Subsection~\ref{sec:testEnvironment} introduces the test environment of the COCO platform. 
Our experiments are described and analyzed in Section~\ref{sec:eval}. First, the experimental setup is defined in Subsection~\ref{sec:setup}, then the results of the runs in the COCO framework, as well as those of the analysis of the multi-objective landscapes, are illustrated and analyzed in Subsection~\ref{sec_results}.
At last, our work is concluded in Section~\ref{sec:concl}.
%-In Sec 3, we detail our approach by
%provides the proofs showing that...

%show that multi-objective problems ...
%-How multiobjectivizing a single-objective optimization problem can remove local optima/how this helps to overcome getting trapped in local optima
%Optimizing the multi-objective problem should reduce local traps that we want to overcome with more information provided by the multi-objective problem. 
%On the generated multi-objective problem we expect MOGSA to perform well due to its ability to slide from one basin of attraction to the next without getting stuck in local optima. 
%As we hope that the arising landscapes contain efficient sets where one is not cut by a ridge indicating a global efficient set, the algorithm should stop there.

\section{Background}\label{sec:background}
To introduce the topic, we will provide definitions for single-objective and multi-objective optimization. 
Furthermore, the basic concept of the multi-objective visualization technique, which is fundamental for our work, is introduced. Afterwards, we will give an overview of related work dealing with multiobjectivization.

In the following, we will study unconstrained single-objective optimization problems of the form:
\begin{equation}\label{eqn1}
\min_{x \in X} f_i(x)
\end{equation}
where $X\subseteq\mathbb{R}^d$ is the set of feasible solutions and $f_i:X\to\mathbb{R}$ is the objective function mapping the $d$-dimensional continuous decision vector $x$ to one function value. If $f(x)$ is to be maximized, it is equivalent to minimizing $-f(x)$. 

%Especially gradient-based algorithms are facing this problem but also sophisticated Evolutionary Algorithms might get trapped in local optima. 

%In continuous optimization, heuristic strategies are often used to search the search space.
%Especially well suited for solving multimodal problems are Evolutionary Algorithms (EA)~\cite{beyer2013ES}.
For solving multimodal problems, Evolutionary Algorithms (EA) are often used~\cite{back1997handbook,beyer2001ES}.
Inspired by the biological evolution, the algorithms use mechanisms such as mutation, recombination, and natural selection in an evolutionary loop for finding an optimal configuration. 
With this strategy, a diverse population is maintained where each individual contains information that can be considered for new search points~\cite{schwefel1993evolution,beyer2001ES}.
%Maintaining a diverse population of solutions is their way of dealing with stagnation.
Further ways of handling multimodal optimization by means of EAs are discussed in depth by Preuss~\cite{preuss2015}.
Due to their population-based nature, EAs are powerful general purpose solvers and therefore very popular in domains like multi-objective optimization~\cite{MOEAs}. 

A multi-objective function $f:X\to\mathbb{R}^p$ with
\begin{equation}
f(x)=(f_1(x),  ..., f_p(x))^T\in\mathbb{R}^p
\end{equation}
is the collection of $p$ single-objective functions as in Eq.~(\ref{eqn1}) with $i=1, ..., p$, which are mapped into a $p$-dimensional continuous objective space. Unlike in single-objective optimization, the aim of multi-objective optimization is to obtain a set of trade-off solutions rather than a single optimal solution. To compare the different solutions, the principle of domination is generally used. To get an understanding of multimodality and the multi-objective version of a local and global optimum in the decision as well as in the objective space, we will provide some definitions as provided in \cite{kerschke2018search,MOlandscapes}.

\begin{defn}
Vector $\textbf{a}=(a_1, ..., a_n)\in\mathbb{R}^n$ \underline{dominates} vector $\textbf{b}=(b_1, ..., b_n)\in\mathbb{R}^n$ (\textbf{a $\prec$ b}), iff $a_i \leq b_i$ for all $i\in{1,...,n}$, including at least one element $k \in {1, ..., n}$ in which $a_k<b_k$. 
\end{defn}

\begin{defn}
A set $A\subseteq\mathbb{R}^n$ is called \underline{connected} if and only if there do not exist two open and disjoint subsets $U_1, U_2\subseteq \mathbb{R}^n$ such that $A \subseteq (U_1 \cup U_2), (U_1 \cap A) \neq \emptyset$, and $(U_2 \cap A) \neq \emptyset$. Further let $B \subseteq \mathbb{R}^n$. A subset $C \subseteq B$ is a \underline{connected component} of $B$ if and only if $C \neq \emptyset$ is connected, and $\nexists D$ with $D\subseteq B$ such that $C\subset D$.
\end{defn}

\begin{defn}
An observation $\textbf{x}\in X$ is called \underline{locally efficient} if there is an open set $U\subseteq \mathbb{R}^d$ with $x \in U$ such that there is no (further) point $\tilde{x} \in (U \cap X)$ that fulfills \textbf{$f(\tilde{x}) \prec f(x)$}. The set of all locally efficient points of $X$ is denoted $X_{LE}$, and each connected component of $X_{LE}$ forms a \underline{local efficient set} (of \textbf{f}).
\end{defn}

For local efficient points in the continuous multi-objective case, Fritz John~\cite{FritzJohn} stated necessary conditions which were extended by Kuhn and Tucker\cite{kuhn1951} to a sufficient condition~. 
Let $x\in X$ be a local efficient point of $X$ and all $p$ single-objective functions of \textbf{f} be continuously differentiable in $\mathbb{R}^n$. 
Then, there is a vector $v\in\mathbb{R}^m$ with $0 \leq v_i \le 1, i=0,...,m$, and $\sum_{i=1}^m v_i=1$, such that

\begin{equation}\label{eqn_fritzJohn}
    \sum \limits_{i=1}^m v_i \nabla f_i(x)=0.
\end{equation}

In case of local efficient points the gradients cancel each other out given a suitable weighting vector $v$.

\begin{defn}
An observation $\textbf{x} \in X$ is said to be \underline{Pareto efficient} or \underline{globally efficient}, if and only if there exists no further observation $\textbf{$\tilde{x}$} \in X$ that dominates \textbf{x}. The set of all global efficient points is denoted $X_E$, and each connected component of $X_E$ is a \underline{(global) efficient set}.
\end{defn}

\begin{defn}
The image (under \textbf{f}) of a problem's local efficient set is called \underline{local (Pareto) front}, whereas the respective image of the union of global efficient sets is called \underline{(global) Pareto front}.
\end{defn}

Since 2016, a Dutch-German research group, whose members are from the Universities of M{\"u}nster (Germany) and Leiden (The Netherlands), has conducted joint research in the field of multi-objective optimization (e.g.,~\cite{kerschke2016towards,MOlandscapes,GrimmeKT2019Multimodality,grimme2019sliding}). They enabled a better understanding of the interactions between the objectives by proposing a way of visualizing the landscapes of continuous multi-objective problems by means of two-dimensional gradient-based heatmaps~\cite{MOlandscapes}. 
Dividing the search space into discrete points, the normalized (approximated) gradients of every objective are aggregated for each grid point. On the path towards the corresponding local efficient set, these multi-objective gradients are accumulated determining the height of the decision vectors and providing information on the closeness of an efficient set. The closer a point is to an efficient point (or set), the smaller is its accumulated value.

These visualizations reveal basins of attraction of local and global efficient sets and ridges in the decision as well as in the objective space.
%As with single-objective problems whose optima fulfill the condition of a zero-gradient in the respective points, the normalized multi-objective gradient is equally zero in local and global efficient sets. 
A basin of attraction is formed by all those points whose multi-objective gradient is directed to the same efficient set. 
When starting in any point belonging to the same basin, one would thus run to the same efficient set by following the multi-objective gradients. 
Due to superposition of basins of attraction, some are cut by other basins resulting in visible ridges.
%In case of local efficient sets, basins of attraction are cut by other basins of attraction due to superposition. 
As it is assumed that ridges only appear for local efficient sets, they could offer a way to escape locality.
Based on these insights, the local optimizer MOGSA was developed exploiting the properties of multimodal multi-objective problems~\cite{GrimmeKT2019Multimodality}. 
By sliding down the multi-objective gradient hill, exploring the sets and jumping across ridges, the algorithm finally finds the global efficient set (see Section~\ref{sec:mogsa} for further details).

In an attempt of utilizing this (beneficial) behavior for single-objective problems, this paper investigates the transformation of originally single-objective into multi-objective problems. 
This process is called multiobjectivization and was introduced by Knowles et al.~\cite{Knowles2001} who differentiated two types of multiobjectivization: the decomposition of the original function into several components, and the consideration of additional objectives to the original function. 
Knowles et al. were the first demonstrating the positive effect of the reduction of local optima in search space by decomposition. 
On two different combinatorial optimization problems (the traveling salesman problem and the hierarchical-if-and-only-if function) the hill climber algorithm based on multiobjectivization outperformed the comparable single-objective optimization algorithm~\cite{Knowles2001}. 

Since then, several authors followed this idea by conducting theoretical and empirical studies on this topic. 
Jensen~\cite{Jensen2004} empirically showed the benefits of multiobjectivization by considering additional objectives that he called \textit{helper-objectives}. These are conflicting with the primary objective but lead to a diversified population which helped to avoid local optima. His experiments reveal significant improvements compared to the average performance of standard genetic algorithms.
For the approach to work, the number of simultaneously used helper-objectives should be low. A high number of helper-objectives in the same run can only be successfully used when changed dynamically.
Positive effects of considering one additional objective are reported by Neumann and Wegener~\cite{neumannWegener2008}. Their theoretical results reveal an improved search behavior of evolutionary algorithms.

That multiobjectivization can have both beneficial and detrimental effects is shown for the decomposition~\cite{Handl2008} as well as for the consideration of additional objectives~\cite{Brockhoff2007}. 
In most research on multiobjectivization, evolutionary algorithms are used for problem solving. 
The main argument in favor of multiobjectivization is, that with a multi-objective instead of a single-objective problem, more information are available that the algorithm can use for improving its search behavior. Contrary to previous research, we will use a local and deterministic multi-objective optimizer to exploit the properties of multi-objective landscapes. The multi-objective setting is generated by considering one additional function.
% a multi-objective problem contains more information than a single-objective problem that in principle can direct the search of evolutionary algorithms

\section{Concept}\label{sec:concept}
Local optima can be deceptive traps for algorithms -- especially when they follow a descent. 
To avoid stagnation and find the global optimum of single-objective problems, we aim at taking advantage of the properties of continuous multimodal multi-objective problems for the single-objective case. 
To this end, we will use the principle of multiobjectivization, which is described in Subsection~\ref{sec:SOtoMO}, followed by a figurative explanation of our approach. 
Next, in Subsection~\ref{sec:MOlandscape}, we will use a simplified example to provide a visual understanding of multi-objective problems and their properties.
The latter are exploited by the local optimizer MOGSA, whose behavior, including the minor modifications made to it, is explained conceptually in Subsection~\ref{sec:mogsa}. 
At last, the test environment used for analyzing our approach is described in Subsection~\ref{sec:testEnvironment}.

\subsection{General Understanding of Multiobjectivization}\label{sec:SOtoMO}
Multiobjectivization is the transformation of an originally single-objective into a multi-objective problem.
In the following, we will explain our procedure and properties of multi-objective landscapes. 

As described above, multiobjectivization can either be achieved by the decomposition of a single-objective problem or by the addition of supplementary objectives.
Within this work, we are interested in the multiobjectivization through the addition of one objective. 
To the initial single-objective optimization problem $f_1$ an additional second single-objective problem $f_2$ will be added. 
Thereby, % to be minimized
a network will emerge, in which all optima of the first objective $f_1$ are connected to all optima of the second problem $f_2$. The arising connections indicate possible areas of efficient sets that can be exploited~\cite{MOlandscapes}.
% Ockham's razor
For not making the landscape more confusing but providing a clear way for the algorithm, we selected a unimodal function for $f_2$.
This way, all efficient sets should point in the direction of the only optimum of the second function which we define as:

\begin{equation}\label{eqn_sphere}
    f_2(x)=\sum \limits_{i=1}^d (x_i-s_i)^2
\end{equation}

in the dimension $d$ with a predefined vector $s\in X$ and $X\subseteq\mathbb{R}^d$. The global optimum of that sphere function is located in $s$.
For the heatmaps and thus for the multi-objective optimizer MOGSA only the gradient direction and the distance to the next efficient set are important. Since the gradient length as such is not relevant, the form of the added function $f_2$ -- meaning whether it is stretched, compressed, moved up or down -- does not make any difference.

An important property of continuous multi-objective problems that we will use in the following was stated in Eq.~(\ref{eqn_fritzJohn}). 
For the bi-objective case this means that the gradients of both objectives become anti-parallel for local efficient sets.
The angle between the two gradients of local efficient points equals $180^\circ$ as the gradients point in two opposing directions and only differ in length.

For the optimization process the second objective does not result in further costs as it is implemented as a constant. This is possible since the function's derivative is known as $\nabla f_2(x)=\sum_{i=1}^d 2\cdot x_i - 2\cdot s_i$.
The additional second objective $f_2$ only serves to create the multi-objective landscape with all its characteristics that can be exploited. 
Traps posed by local optima in the single-objective case are replaced by efficient sets that should guide the search to the global efficient set in the multi-objective setting.
To ensure visualization we set the dimension $d=2$ for all considered problems.

\subsection{Multiobjectivization and the Resulting Landscape}\label{sec:MOlandscape}
\begin{figure}[b!]
	\includegraphics[width=0.475\textwidth, trim = 25mm 9mm 9mm 9mm, clip]{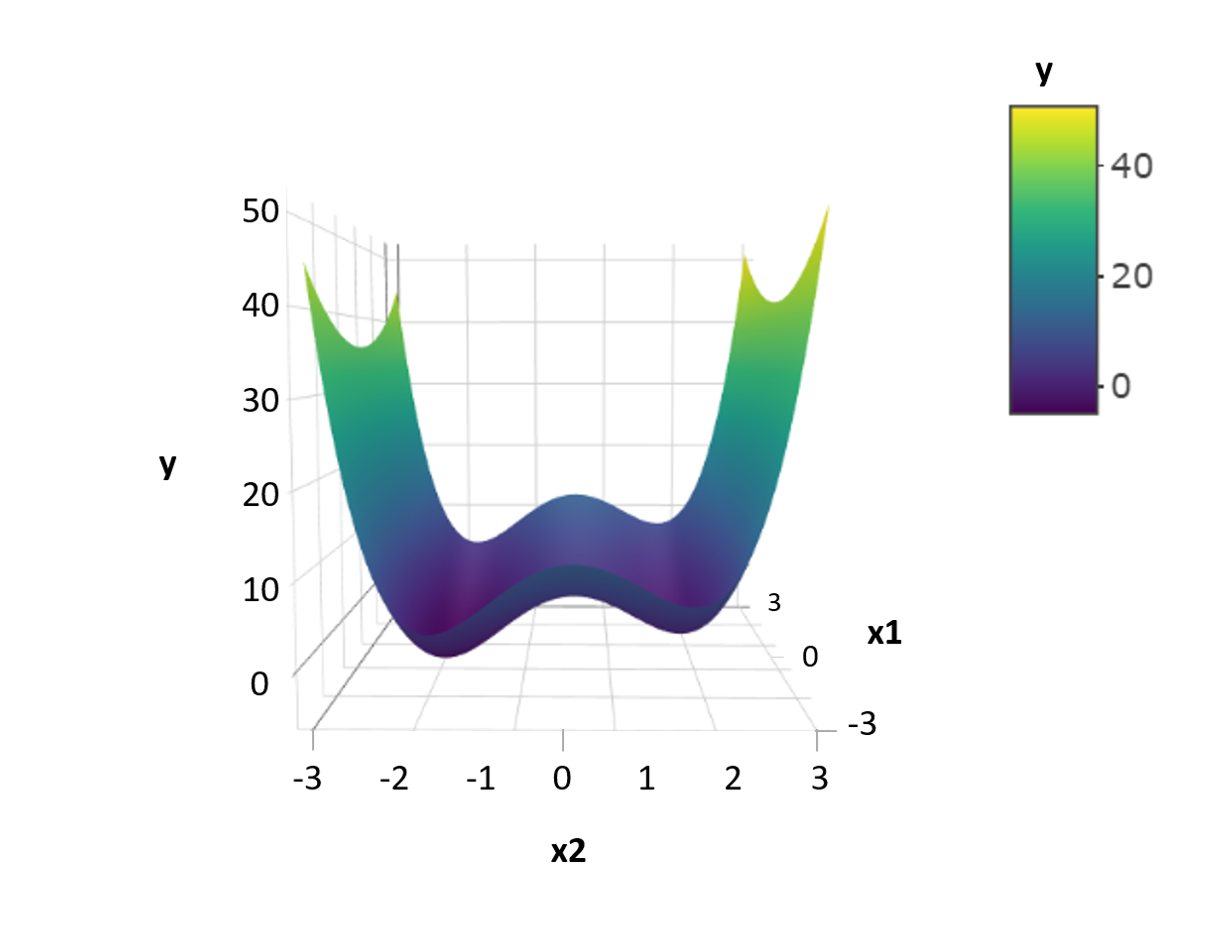}
	\centering
	\caption{\label{fig:sop2} A single-objective function $y=f_1(x)$ with one local (right) and one global optimum (left). The sidebar represents the mapping of the function value $y$ to the used coloring.}
\end{figure}

In the following, we will provide a figurative description of the process of multi\-objectivization and the properties of the resulting multi-objective problems. 
The principle of our procedure will be illustrated by a rather simple scenario with the initial single-objective function $y=f_1(x)=x_1^4-5\cdot x_1^2+x_1+x_2^2+3$.
Figure~\ref{fig:sop2} depicts this multimodal single-objective problem with its two peaks in a three-dimensional plot. 
With a function value of $-4.86$ the global minimum is located in $(-1.63, 0)$, the local one with a fitness of $-1.69$ in $(1.53, 0)$.
Imagining the behavior of a basic gradient descent algorithm starting in $(2.5, 2)$, it is obvious that the algorithm would converge to the local optimum by following the negated gradients. 
There, it would finally get stuck as no step in any direction would bring an improvement. 
The algorithm stops without noticing that there are other optima or rather that the found solution is not the best.

\begin{figure}[t]
    \centering
    \includegraphics[width=0.6\textwidth, trim = 4mm 5mm 14mm 5mm, clip]{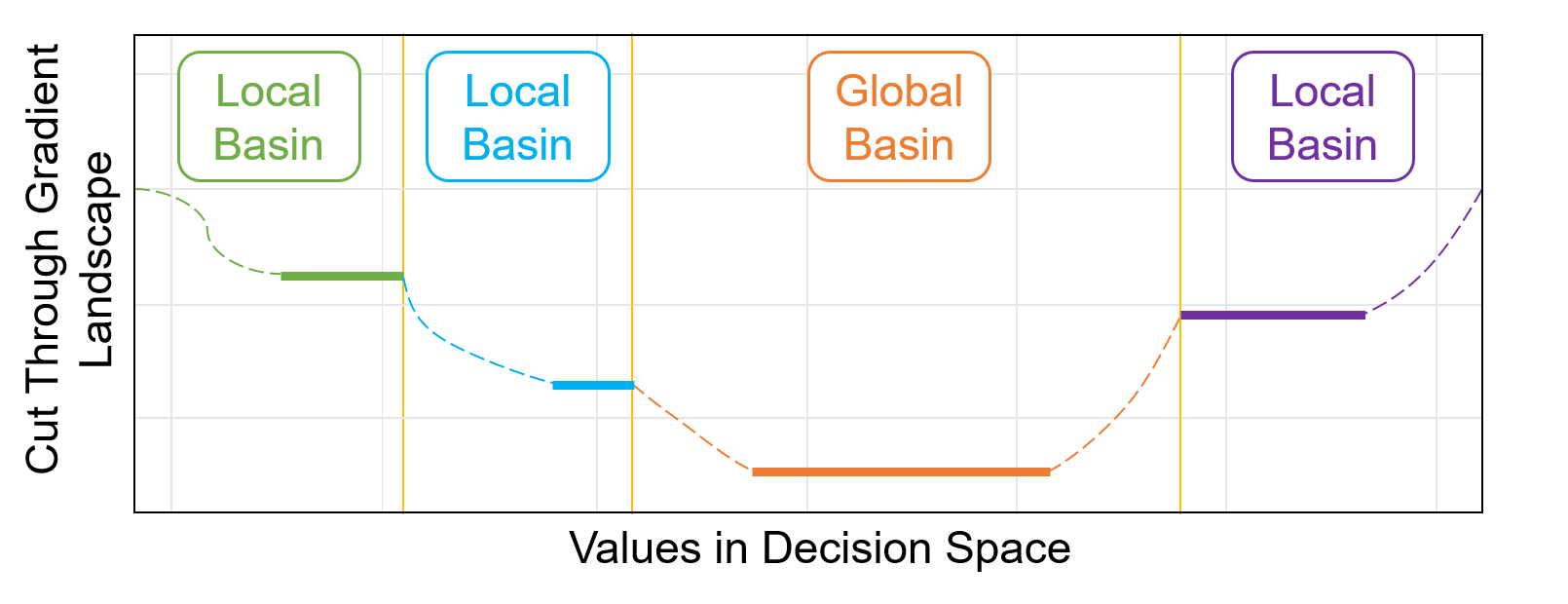}
    \caption{Schematic depiction of the superposed structure of basins of attraction (areas between the vertical red lines). The colored horizontal lines display the efficient sets of the respective basin and the dotted lines the path towards them. Reprinted from~\cite{GrimmeKT2019Multimodality}}
    \label{fig:basinCut}
\end{figure}

To avoid such a stagnation, the single-objective problem will be transferred into a multi-objective problem by adding an additional function $f_2$. 
By visualizing the created bi-objective problem in a gradient-based heatmap, the efficient sets become visible as well as their respective basins of attraction. 
Due to the interaction effects between the objectives, the basins of attraction superpose each other resulting in visible ridges along the borders of the basins. 
The superposed structure is schematically depicted in Figure~\ref{fig:basinCut}. The basins of attraction can be regarded as funnels towards their respective efficient set. 
Picturing a ball following the multi-objective gradients, it would roll down the dotted path towards the efficient set that is displayed by the colored horizontal lines. 
Ridges, illustrated by the vertical red lines, cut efficient sets due to the superposition of basins of attraction. 
All efficient sets that are cut by a ridge are assumed to be locally efficient. Consequently, global efficient sets should be ridge-free~\cite{GrimmeKT2019Multimodality}.
By following the multi-objective gradients and walking along the efficient sets to the next basin of attraction one should find the global efficient set.

The global optimum of both single-objective problems $f_1$ and $f_2$ is equally globally efficient from a multi-objective point of view. No improvement in the dimension of the respective objective is possible.
Therefore, we aim at finding the global efficient set that leads us to the global optimum of the initial single-objective problem $f_1$.
For this purpose, we use the local multi-objective optimizer MOGSA that explores the efficient sets, jumps over the ridges into other basins of attraction until the global efficient set without any ridges is found. 

\begin{figure}[t]
  \begin{subfigure}[b]{0.475\textwidth}
    \includegraphics[width=\textwidth]{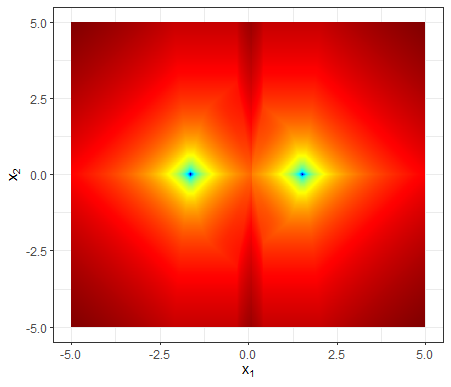}
    %\caption{single-objective problem }
    \label{fig:heatSOP}
  \end{subfigure}
  \hfill
  \begin{subfigure}[b]{0.475\textwidth}
    \includegraphics[width=\textwidth]{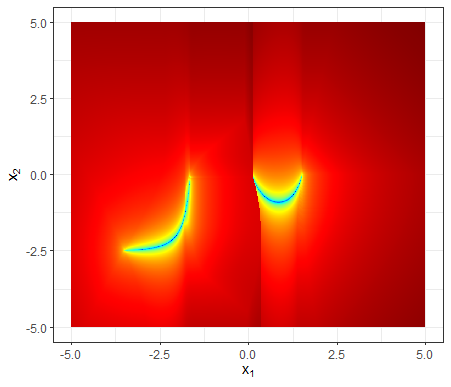}
   % \caption{multi-objective problem, joint visualization of an unimodal problem and a multimodal (two optima)}
    \label{fig:heatMOP}
  \end{subfigure}
\caption{\label{fig:heats} Depiction of two gradient-based heatmaps. On the left, the decision space of the already introduced bimodal single-objective problem is displayed. Adding a sphere function to this problem making it multi-objective results in the decision space of the right image.}
\end{figure}

The transformation of a single-objective into a multi-objective problem is visually illustrated in Figure \ref{fig:heats}. On the left side, the decision space of the single-objective problem from Figure \ref{fig:sop2} is depicted in a heatmap. 
The different colors denote the distance -- w.r.t.~the gradient descent -- to the efficient set of the respective basin of attraction. The darker the red, the further away is the point. The efficient sets -- here, the two optima since this is the single-objective case -- are colored in blue, usually surrounded by an area of green to yellow. 

Adding the additional function $f_2$ from Eq.~(\ref{eqn_sphere}), with its optimum in $s=(-3.5, -2.5)$, to the previously discussed scenario results in the heatmap on the right side. Now, the efficient sets and their basins of attraction become visible.
A ridge cuts the local efficient set (the set on the right side of the image) while the global efficient set on the left side is ridge-free. 
At each ridge-free end of the efficient sets an optimum of one of the two functions is located.

\subsection{Search Behavior and Implementation of MOGSA}\label{sec:mogsa}
Ridges help the optimizer MOGSA to escape local efficient sets. %%%%
As proposed by Grimme and Kerschke~\cite{GrimmeKT2019Multimodality}, the algorithm takes advantage of the properties of the multi-objective problems' landscapes in two repeating phases.
Following the problem's multi-objective gradients in the first phase, MOGSA finds an efficient set that can be explored in the second one. 
We will make use of the local optimizer with slight adaptions to support our approach.
First, we will describe the search behavior of MOGSA in a simplified setting, supported by a schematic illustration. % the way MOGSA would take in a simplified setting
Afterwards, the two phases and our changes made are described and explained in detail.
%A schematic illustration in Figure \ref{fig:wayMOGSA} shows the way MOGSA would take in a simplified setting of two basins of attraction.

\begin{figure}[t!]
	\includegraphics[width=0.6\textwidth]{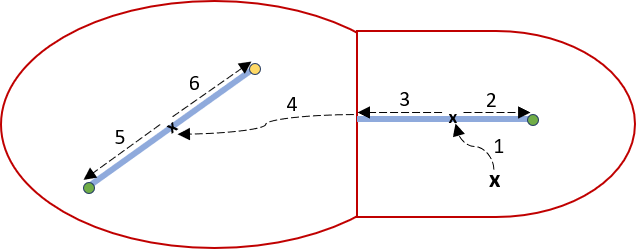}
	\centering
	\caption{\label{fig:wayMOGSA} Schematic view on the decision space of a bi-objective problem with two basins of attraction (encircled in red). The optima of both single-objective functions are indicated by dots (green for the first, yellow for the second objective) being located on two efficient sets (blue lines). The dotted arrows display the chronological search behavior of MOGSA starting in point x.}
\end{figure}

\paragraph{Way of MOGSA:} %Search Behavior
The interaction of the two repeating phases is schematically depicted in Figure \ref{fig:wayMOGSA}. 
The presented bi-objective problem is comprised of one bimodal function $f_1$ and one unimodal function $f_2$. 
The optima are indicated by dots (green for the first, yellow for the second objective). 
The blue lines illustrate the efficient sets which are located in different basins of attraction (area within the red borders). 
Being cut by a ridge, the efficient set in the right basin of attraction represents the local efficient set. The global efficient set and thereby the global optima of both single-objective functions are located in the left basin of attraction.\\
Starting in point x, MOGSA executes the first phase by following the multi-objective gradient to find a point on the local efficient set in the respective basin (1). From there, the single-objective gradient of the first function $f_1$ is followed until the (green) optimum is reached (2). Going back to the first found point on that efficient set, the exploration phase is repeated for the second objective $f_2$ until the end of the set is reached and a ridge found (3). With this, the second phase stops and the first one is started again searching for the efficient set in the new basin of attraction (4). Once the set has been reached, $f_1$ is followed until the (global) optimum is found (5). Eventually, the algorithm would jump over it in a first try but with a local search, the optimum would be found. Next, $f_2$ is followed from the first found point on that set (6). As both sides of the efficient set are ridge-free, MOGSA stops successfully with the global efficient set and thus the global optimum of the first function found.

\paragraph{First Phase:}
% approximating the gradient (for each gradient approximation four function evaluations in the two dimensional case)
The aim of this phase is to find an efficient set. As a first step, the multi-objective gradient is computed for the current position by combining the normalized (approximated) single-objective gradients.
The combined gradient contains information about the direction to the attracting efficient set.
In the bi-objective case, two single-objective gradients pointing in the same direction result in a large multi-objective gradient, i.e., a length close to two. 
A small multi-objective gradient or rather a zero-gradient on the other hand means that the single-objective gradients point in opposite directions. 
In this case, an efficient set is reached and the local search stops. Otherwise, a scaled step in the direction of the multi-objective gradient's descent is taken. 
If the step leads to a crossing of the boundaries, the algorithm is placed back on the bounds. When the algorithm would have to be placed at the same position a second time, it is randomly restarted within the feasible area. 
%Grimme and Kerschke proposed a restart using Optimal Augmented Latin Hypercube Sampling in case of a step that is too small. However, this was not implemented as it is not necessary for our approach. 
%Placing the minimum of the second objective within the boundaries, the added function guides the algorithm away from the bounds where a scenario requiring such a procedure could occur.
Gradient-descent steps are performed until an efficient set is reached (or crossed). If the set has been crossed, the last two consecutive points are located on opposite sides of the attracting efficient set. 
As their multi-objective gradients are both pointing towards the efficient set, they are pointing in nearly opposing directions. Thus, the angle between them will be large. 
If the angle is bigger than $90^\circ$, an interval bisection procedure weighted by the ratio of the lengths of the two gradients is performed. This way, a point on the efficient set will be found.

\paragraph{Second Phase:}
Having found an efficient set, it will be explored in the second phase. From the locally efficient point that was found in the first phase, MOGSA starts with following the normalized and scaled gradient of the first objective $f_1$ as long as the step size is not too small. 
As in the first phase, a possible leaving of the feasible area is checked. This would result in following the second objective $f_2$ from the initially found efficient point of that set.
However, when staying within the bounds, MOGSA continues to follow the single-objective gradient of $f_1$ until one of the following three scenarios occurs: the algorithm could have landed directly in the optimum, passed an optimum, or left the efficient set due to a ridge.\\
A length of (nearly) zero for the single-objective gradient indicates that the algorithm has found the single-objective optimum of $f_1$. 
As a next step, the gradient of $f_2$ would be followed to check whether the currently exploring efficient set could be globally efficient. This is likely if the set is not cut by a ridge.\\
In the other two scenarios, the efficient set was left. 
When the last point on the efficient set $x^{(t)}$ and the point after the step taken $x^{(t+1)}$ are still in the same basin of attraction, MOGSA has passed an optimum or the optimum is located close to $x^{(t+1)}$. 
This is indicated by an angle of bigger than $90^\circ$ between the single-objective gradient of $x^{(t)}$ and the one of $x^{(t+1)}$. Both gradients are pointing to the same optimum but from different sides.
To find the local single-objective optimum of $f_1$, the Nelder-Mead algorithm is used, originally published in 1965 by John Nelder and Roger Mead~\cite{nelder1965simplex}.
The direct search method does not require any derivative information and uses the concept of a simplex. 
Instead of a single starting point, $d+1$ vertices are used for $d$-dimensions. In every iteration, one vertex is exchanged by a new point.
Since we are only interested in the optima of the first function, the local search was added to the proposed algorithm and is solely performed in the direction of $f_1$.
Once the optimum has been found, the second phase starts again from the point that was found first on the set. This time, the second objective's gradient is followed.\\
In the event of having left the basin of attraction, the algorithm has crossed a ridge to an adjacent basin. 
For a point on the efficient set, the angle between the two single-objective gradients is $180^\circ$.
Leaving the basin of attraction leads to a new attracting efficient set. The currently followed single-objective gradient should still point to the same optimum and thus in the same direction as before. For the gradient of the other function on the other hand, the attracting optimum has changed by entering the new basin of attraction. In most cases, the angle between the two single-objective gradients of $x^{(t+1)}$ is smaller than $90^\circ$. 
The existence of a ridge indicates a local efficient set that we are not looking for. Thus, MOGSA will not start exploring the previous efficient set in the direction of the second objective but stay in the newly found basin. There, the first phase is started again to find a new efficient set.

The two phases are repeated until a set without any ridges and thus a probable global efficient set is found. 
Each point the algorithm considers a probable optimum is saved and evaluated in the end. 
For preventing MOGSA from getting trapped in an endless loop between different efficient sets (e.g., from efficient set a into efficient set b, back to a and again to b), every probable optimum is compared to all the already visited optima. 
In case the algorithm runs into the same optimum again, a random point within the given bounds becomes the starting point for the first phase.

\subsection{Test Environment}\label{sec:testEnvironment}
As already stated within the introduction, our aim is to analyze how the properties of multi-objective landscapes can be exploited for single-objective optimization. We want to explore whether the multi-objective optimizer MOGSA is able to find the global optimum of a single-objective problem that is transferred to a bi-objective problem through multiobjectivization.
For testing the search behavior on many single-objective problems with different characteristics and difficulties, we use the benchmark platform COCO~\cite{hansen2016cocoplat}. 
In this section we will introduce the platform and the performance measure used for our experiments.

The COCO platform is used for comparing continuous optimizers in a black-box setting. For this, an experimental framework as well as post-processing facilities are provided.
Different test suites contain benchmark functions for single-objective optimization with and without noise as well as noiseless bi-objective problems. We use the 24 noiseless single-objective test functions of the Black-Box Optimization Benchmark (BBOB)~\cite{Definition_noislessFunctions}.
The benchmark functions are categorized in five subgroups according to similar properties such as multimodality, ill-conditioning, and non-separability of the variables. %and non-linearities.
Furthermore, different instances, e.g., shifted, rotated and stretched versions of each problem are available. 
For each of the 24 problems of the BBOB test suite multiple instances are provided.
This way, an algorithm can be exposed to a variety of function characteristics. 
A problem or problem instance triplet $p^3=(d, f_{\theta}, \theta_i)$ is formally defined by the search space dimension $d$, the objective function $f$ to be minimized and its instance parameters $\theta_i$ of instance $i$. 
A set of parametrized benchmark functions $f_\theta :\mathbb{R}^d \rightarrow \mathbb{R}^m, \theta \in \Theta$ and the corresponding problems $p^3$ are considered.
For performance assessment, it is aggregated over all $\theta_i$-values.

For our experiments, we consider the 24 problems from the BBOB suite as potential functions of interest ($f_1$), which we aim to optimize.
The multiobjectivization process is part of the algorithm and thus independent of the first objective. 
The added second function $f_2$ will not be changed, neither within the optimization process of a problem nor for optimizing other functions. The additional objective can be regarded as a constant. 
Thus, COCO only considers the first function. Note that the usage of the second objective does not generate any further costs in terms of function evaluations -- as $f_2$ as well as its gradient $\nabla f_2$ are known. 

Denoted as the runtime, the number of function evaluations conducted on a given problem is the central performance measure of COCO.
When the algorithm reaches or surpasses a prescribed target function value $f_{target}$, the problem is considered as being solved regardless of whether the algorithm continues running or not. 
Function evaluations conducted after having solved the problem are not recorded anymore. 
The target function value allows small deviations ($\Delta f$) from the best value possible to reach ($f_{opt}$) and is thus defined as $f_{target}=f_{opt}+\Delta f$. 
The optimal function value is defined for each instance of the benchmark problems individually. The smallest considered precision to reach is $10^{-8}$~\cite{experimentalSetup}.
For the experiments, different target precisions are defined. 
From each run on the problem instance triple $p^3$ runtime measurements are obtained for each target value reached in this run.

Ways for aggregating the resulting runtimes are the average runtimes (aRT) and the empirical (cumulated) distribution function (ECDF) of runtimes~\cite{hansen2016cocoplat}. 
As we are primarily interested in the success of the algorithm and the precision of the found optimum, we will focus on the ECDF that displays the proportion of problems solved within a specified budget. 
The runtime distributions can be displayed for each problem individually but also aggregated over the function groups or over all functions. %The performance of the algorithm on specific function instances is not presented.

\section{Evaluation}\label{sec:eval}
This section deals with the realization and the analysis of our concept with the aim of reducing local traps for single-objective problems by multiobjectivization.
To conduct experiments on several single-objective functions and assess the performance of the two local optimizers Nelder-Mead and MOGSA, the COCO framework is used.
Furthermore, a visualization technique based on multi-objective gradients offers insights into the landscape of the multi-objective problems. 

First, the setup for experiments conducted by the COCO framework is described. 
Then, the results of this experiment, the visual inspection of the landscape, and the behavior of MOGSA are described and analyzed.

%- Results from a visual inspection of the landscapes and the behavior of MOGSA
%- analyzing the heatmaps/landscapes of the multi-objectivized SOPs and the respective behavior of the optimizer MOGSA - is he able to find the global efficient set of the single-objective problem?
%- comparison of local optimizer Nelder-Mead that was run in a single-objective setting and local MOGSA that was run in a multi-objective setting (both run on the same functions)

\subsection{Experimental Setup}\label{sec:setup}
The performance of MOGSA on the initial single-objective problem is assessed by the COCO framework~\cite{hansen2016cocoplat}. 
To evaluate the transformation of adding an additional function, the local multi-objective optimizer MOGSA is compared to the local single-objective optimizer Nelder-Mead on a testbed with several different optimization problems.
In this Section, the characteristics of the machine and the environment used for generating the results of the comparison are provided.

The tests have been run on a Windows 10 Education (2018) computer with AMD Ryzen 5 3500U at 2.10 GHz with 8 GB RAM. 
Python 2.7.17 (64 bit) was used to run MOGSA and the test environment that is provided by the COCO framework. 
The simplex algorithm Nelder-Mead was provided by the package skipy.optimize of Python~\cite{scipy,2020SciPy-NMeth}. 

A detailed description of the experimental setup and of how the results are displayed with COCO can be found in~\cite{experimentalSetup}. The Nelder-Mead algorithm and MOGSA have been run on each problem of the noiseless BBOB test suite. The suite contains continuous single-objective benchmark functions with different properties. A documentation of the test functions can be found in~\cite{Definition_noislessFunctions}.
For each of the 24 problems 15 different function instances (Ntrail) were considered (and optimized). The performance is evaluated over all Ntrial trials. 
For both algorithms, nothing is known about the system which has to be optimized apart from the search domain $(-5, 5)^d$ and the starting point. In our experiments, we define the search space dimension $d=2$. 
To be able to evaluate the overall behavior of both algorithms, each started from ten different points within the search domain: $(5,5)$, $(-5,5)$, $(5,-5)$, $(-5,-5)$, $(2,4)$, $(4,2)$, $(-2.5,4)$, $(1,-2)$, $(0,0)$, $(-4.5,0)$. 
These points were manually spread over the search space and are intended to generate results that are independent of possible advantages in functional structures.\\
For each of the ten runs on all function instances of the test suite the same parameter setting, the same initialization procedure, the same budget and the same termination criteria were used. The maximum budget for evaluations done by an optimization algorithm is \textit{dimension$\times$budget-multiplier} with a fixed budget-multiplier of 100 for our experiments.

The number and the initialization of the variables for MOGSA were set as proposed in~\cite{GrimmeKT2019Multimodality}. Only the scaling factors $\sigma_1$ and $\sigma_2$ for the step size for both of the two phases of MOGSA were changed. A step size of one would be too big so that the algorithm would not be able to identify the structure of the problems. Instead, the algorithm would leave basins of attraction without finding the efficient set. Thus, we defined $\sigma_1=\sigma_2=0.1$. As the stopping criterion we implemented the variable \textit{steps}. Running the first phase of finding an efficient set counts as one step. Internally, this phase stops when an efficient set is found or more than $1,\!000$ new x-values were generated while following the multi-objective gradient. For phase two, every movement from one point $x_t$ to another point $x_{t+1}$ for following $f_1$ or $f_2$ counts as one step. When MOGSA has not found an efficient set without any ridge within 1000 steps, the algorithm stops. The transformation of a single-objective benchmark problem into a multi-objective one is internally performed by MOGSA.
As in the introductory problem in Figure~\ref{fig:heats}, we defined $s=(-3.5, -2.5)$ for the additional second objective $f_2$ from Eq.~(\ref{eqn_sphere}).

For visualizing the generated problems as gradient-based heatmaps, we used R version 4.0~\cite{R}, as well as the R-packages moPLOT~\cite{SchaepermeierGK2020}, ggplot2~\cite{gglot2} and smoof~\cite{smoof}.

\subsection{Experimental Results}\label{sec_results}
From each of the ten starting points, experiments were conducted for the simplex algorithm Nelder-Mead and the multi-objective optimizer MOGSA. Both algorithms were run on 15 instances of each of the 24 functions $(f_1-f_{24})$ of the COCO framework with the previously described setup. 
For MOGSA, a fixed additional second objective was added to create a multi-objective landscape.
The generated results are discussed and compared in the following.
Furthermore, individual problems are illustrated in gradient-based heatmaps and visually ana\-lyzed.
% Afterwards

\subsubsection{Performance Assessment}
First, we will look at the runtime distributions aggregated over all 24 functions of the BBOB test suite and explain the plots which resulted from the experiments.
Afterwards, we will focus on specific functions.
% take a closer look at

\begin{figure}[b!]
	\centering
	\includegraphics[width=0.485\textwidth]{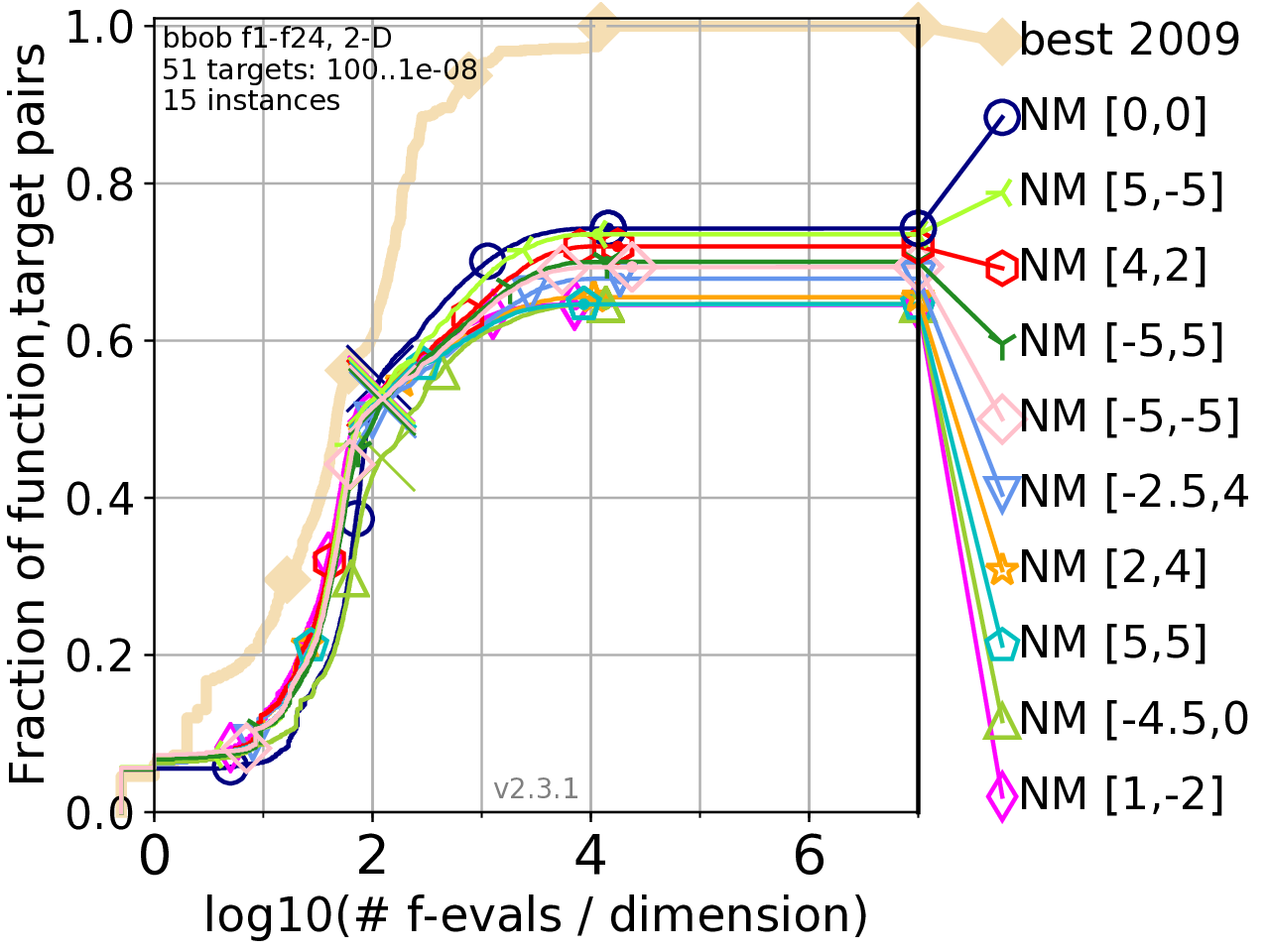}
%	\caption{\label{fig:coco_nelder} ECDF of the runtime of ten runs of Nelder-Mead (each from a different starting point) in dimension 2 over 51 targets. The ECDF is aggregated over all functions of the BBOB test suite for each run.
	%The ECDF of runtimes on all single-objective functions of the BBOB framework for ten runs of the Nelder-Mead algorithm.
%	}
%\end{figure}
\hfill
%\begin{figure}[h!]
%	\centering
	\includegraphics[width=0.485\textwidth]{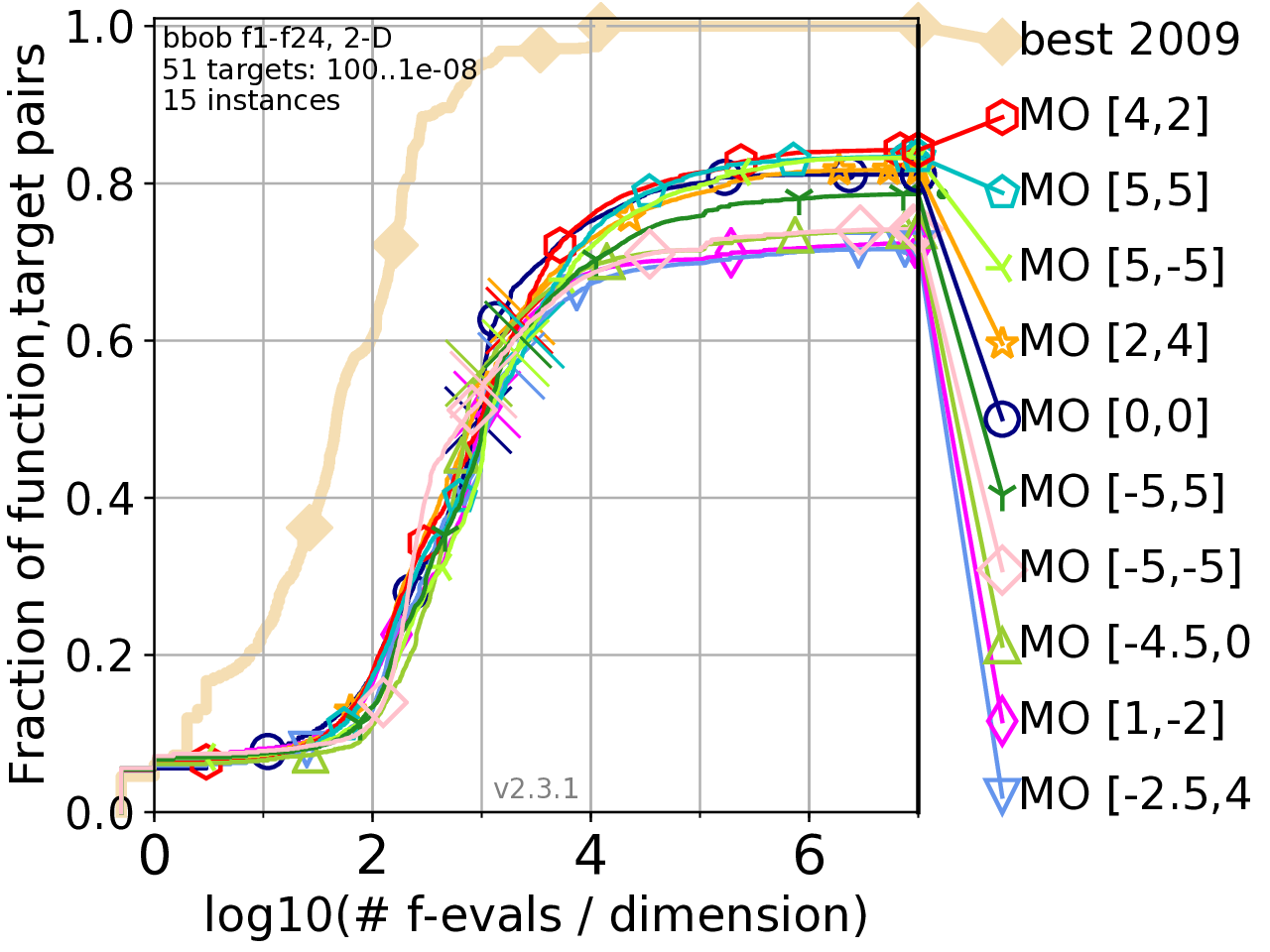}
	\caption{\label{fig:coco_mogsa} ECDF of the runtime of ten runs of Nelder-Mead (left) and MOGSA (right) in dimension 2 over 51 targets. Each run was started in a different starting point. The ECDF is aggregated over all functions of the BBOB test suite for each run.
	%Aggregated over all functions of the BBOB test suite, each run obtains one ECDF.
	}
\end{figure}

In Figure~\ref{fig:coco_mogsa}, the empirical cumulated distribution functions (ECDF) of runtimes of Nelder-Mead (left) and MOGSA (right) have been depicted. 
Both plots show the ECDF of runtimes on a set of problems formed by $24\times 15$ function instances in dimension $d=2$, each with 51 target precisions between $10^2$ and $10^{-8}$ uniform on a log-scale.
The precision is displayed on the y-axis. A value of $0.0$ corresponds to a precision of $10^2$, a value of $1.0$ to a precision of $10^{-8}$.
On the x-axis, the number of objective function evaluations divided by the dimension $d$ is represented on a log-scale.
Each colored line in the plot represents one of the ten runs whose starting points are indicated next to the abbreviated name of the algorithm (NM for Nelder-Mead, MO for MOGSA). 
The light thick line portrays the artificial best algorithm of BBOB-2009 as reference algorithm.
The crosses on the plots illustrate the median of the maximal length of the unsuccessful runs to solve the problems aggregated within the ECDF.

The ECDF can be read in two different ways when regarding one of the axes as the independent variable and the other one as the fixed.
In the plot of Nelder Mead, for example, we can find the number of function evaluations (x-axis) necessary for all runs to solve 40\% of the problem-target combinations (fixed y-axis). Solving 40\% of the problems (on average) corresponds to reaching a target precision of $10^{-2}$. That precision was reached by all of the ten runs within $10^2\cdot d = 200$ function evaluations (value of 2 on the x-axis).

Comparing the two plots, we observe that aggregated over all target-problem combinations, MOGSA obtains better results than Nelder-Mead concerning the final reached precision. 
All runs of MOGSA solved at least 70\% of the target-problem combinations after $10^7\cdot 2$ function evaluations and thus achieved an average precision of at least $10^{-5}$.
For Nelder-Mead this is the best achieved value of all the runs.
In comparison, the best run of MOGSA even solved 85\% of the problems.
With a small number of function evaluations, however, Nelder-Mead finds points with better function values. Only from a number of $10^4\cdot 2$ function evaluations on, MOGSA achieves better results.

These plots of the aggregated ECDF over all functions indicate a success for our procedure. However, we need to examine the functions separately as each possesses different characteristics.
Besides, in the set of benchmark functions are some unimodal problems. These are not of our interest as we aim to overcome local traps which do not exist for problems with only one optimum. 
Instead of directly running to that optimum, MOGSA would find the efficient set first to explore it. Although it finds the optimum -- with the same precision as Nelder-Mead -- it takes more function evaluations. 
The only unimodal function on which Nelder-Mead performed better than MOGSA concerning the target precision is the Step Ellipsoidal function $f_7$. 
However, apart from a small area close to the global optimum the gradient is zero almost everywhere.
As MOGSA, contrary to Nelder-Mead, uses the approximated gradients for its search, this result is not surprising. 

\begin{figure}[p]
    \centering
    \includegraphics[width=0.43\textwidth]{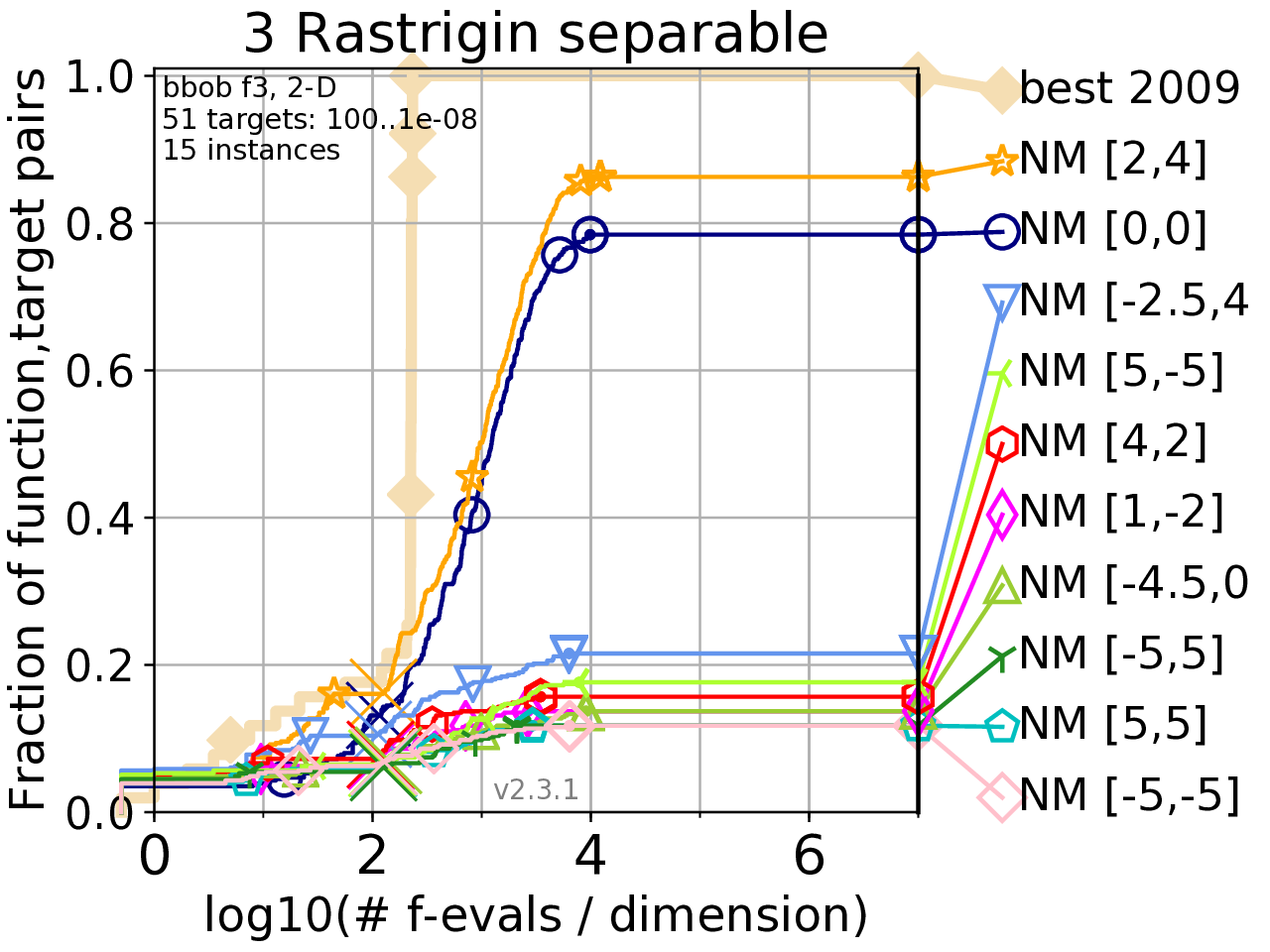}\hfill
    \includegraphics[width=0.43\textwidth]{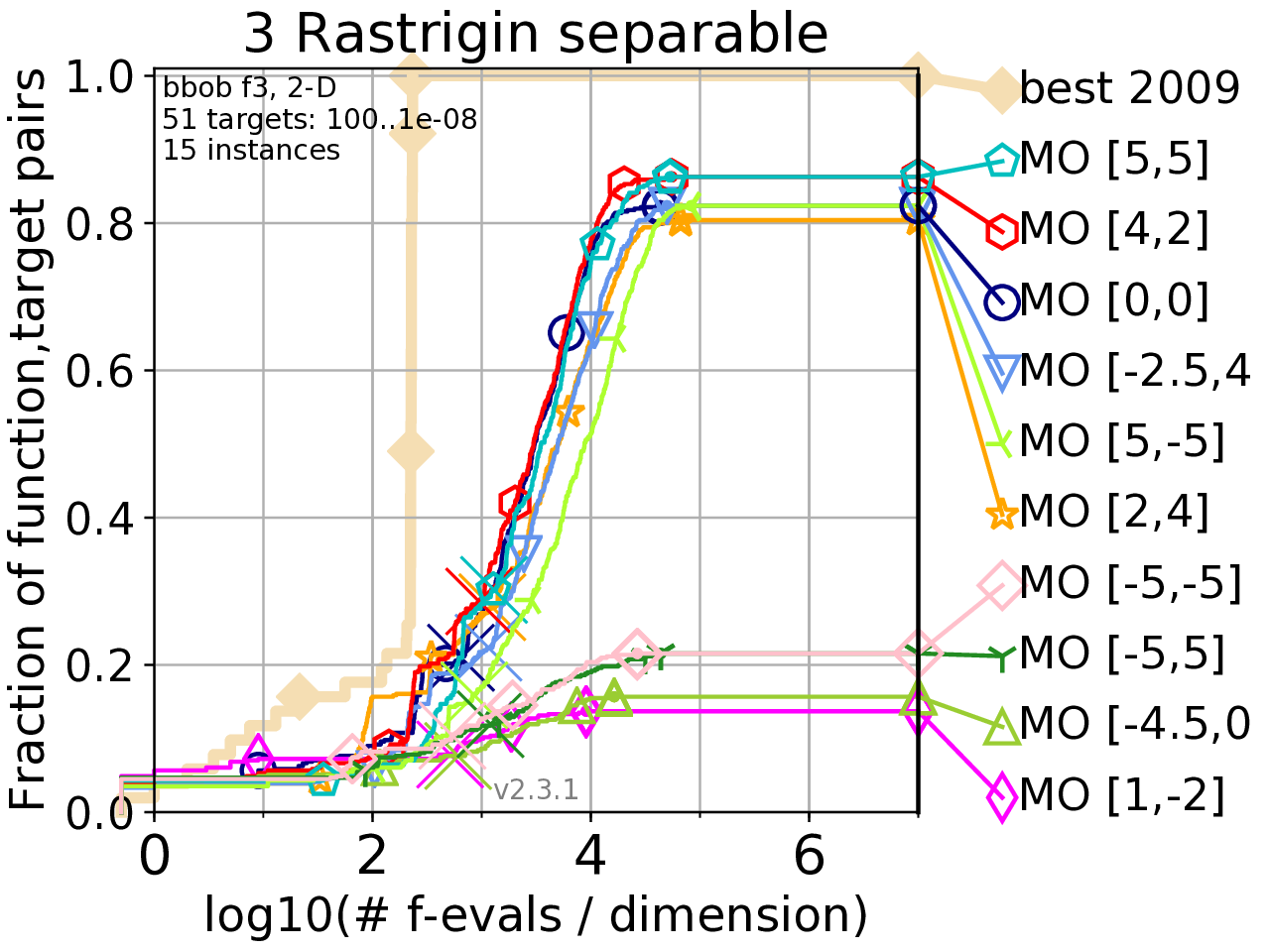}\\[0.25em]
    \includegraphics[width=0.43\textwidth]{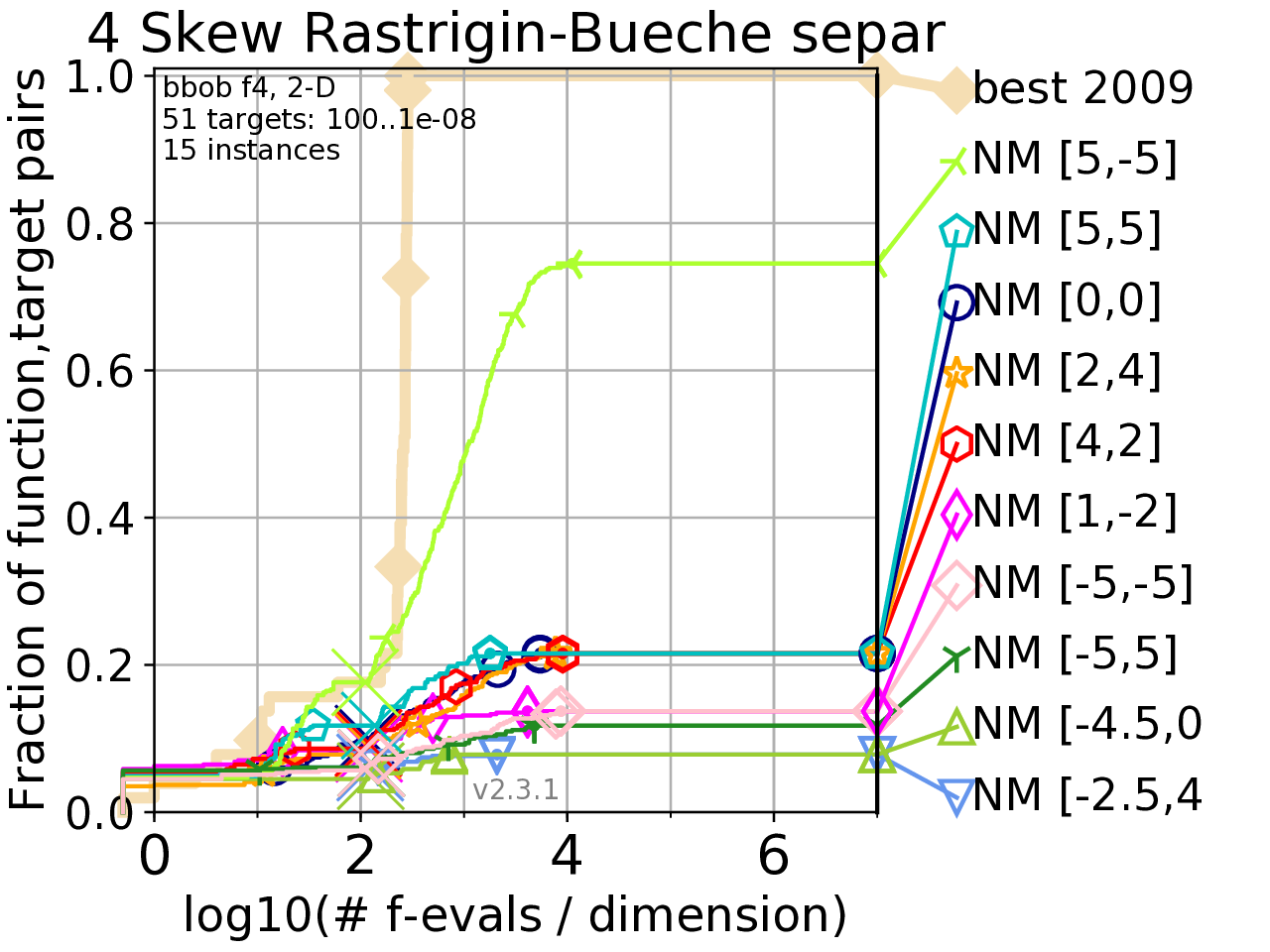}\hfill
    \includegraphics[width=0.43\textwidth]{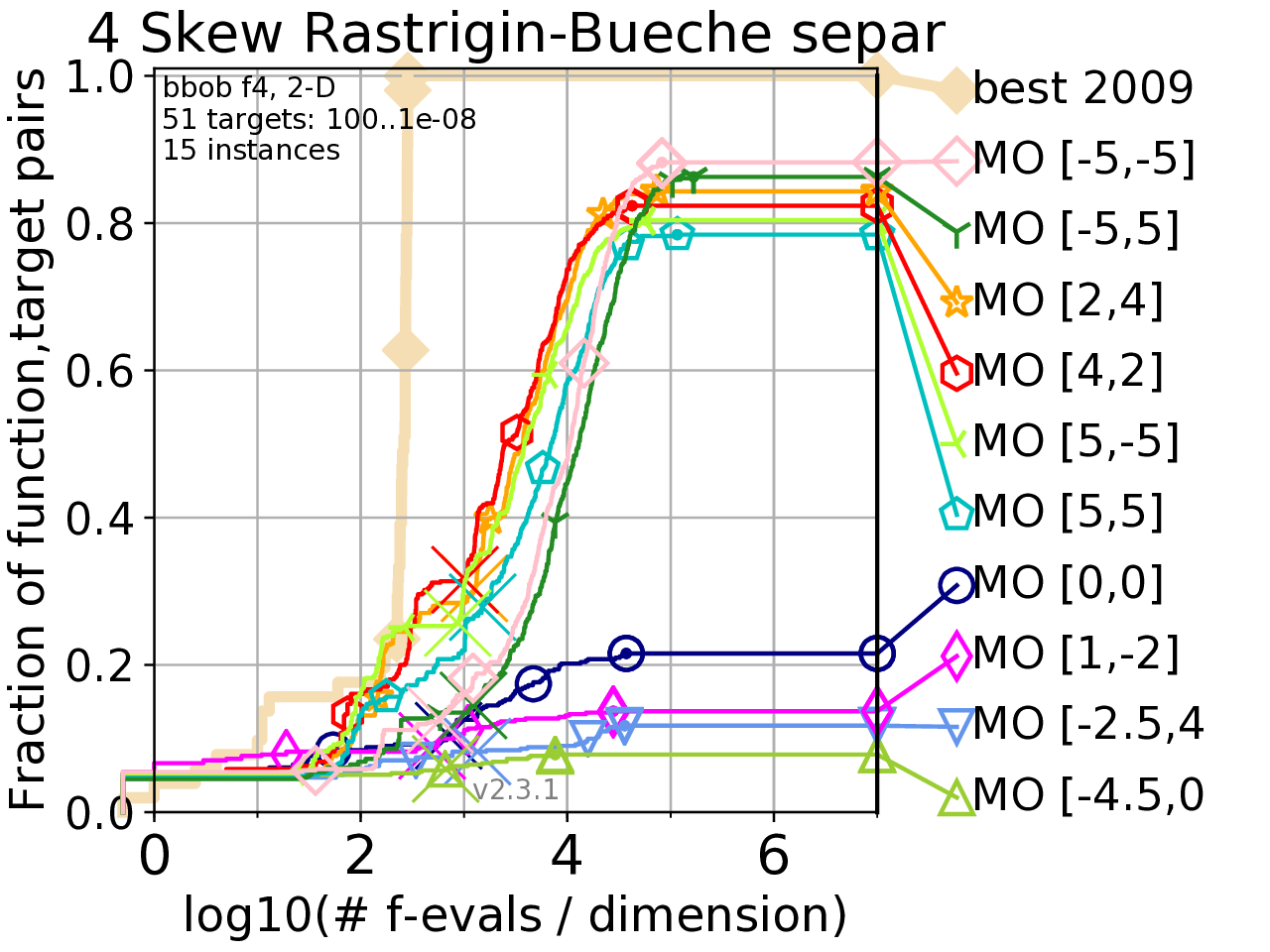}\\[0.25em]
    \includegraphics[width=0.43\textwidth]{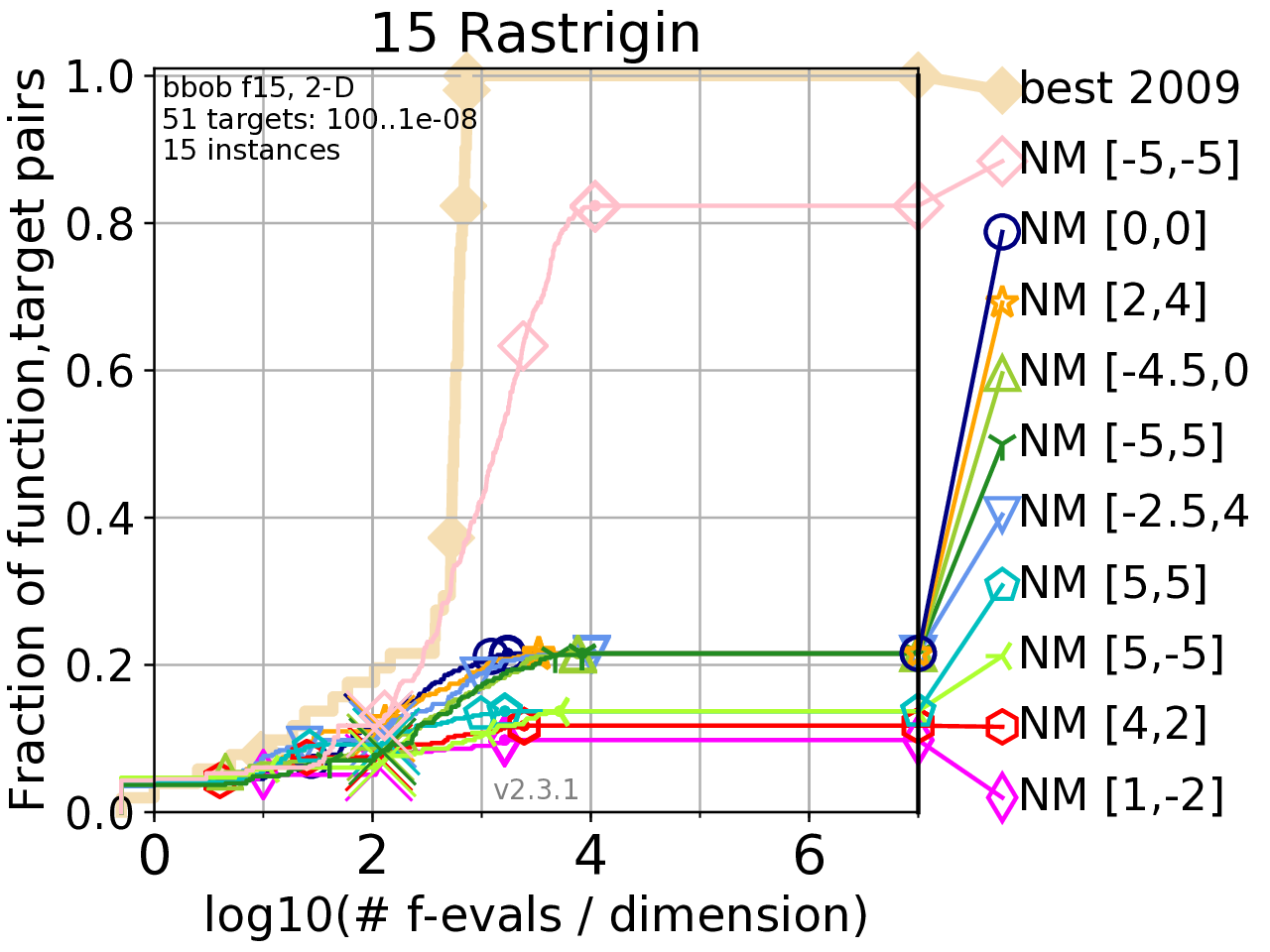}\hfill
    \includegraphics[width=0.43\textwidth]{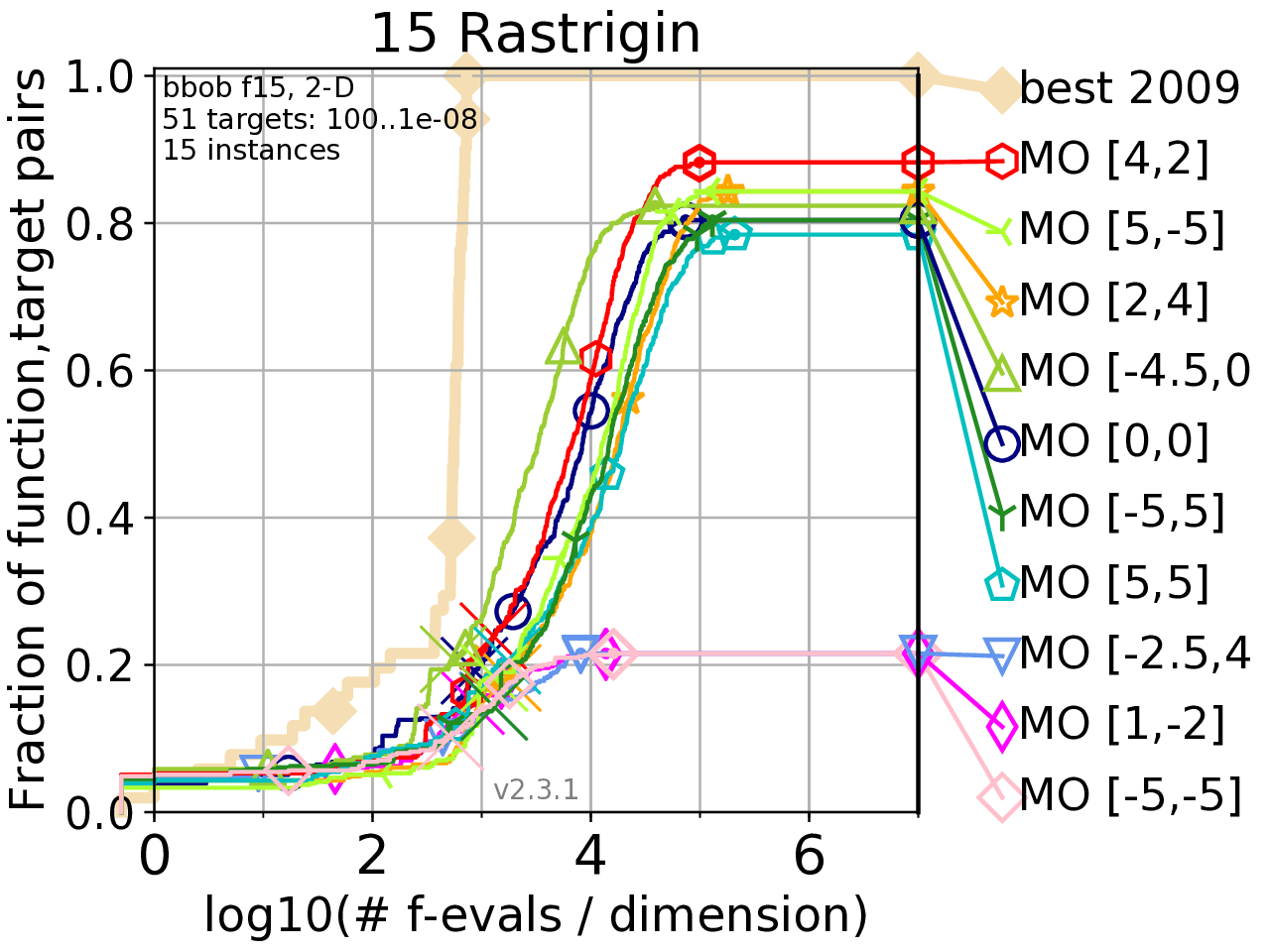}\\[0.25em]
    \includegraphics[width=0.43\textwidth]{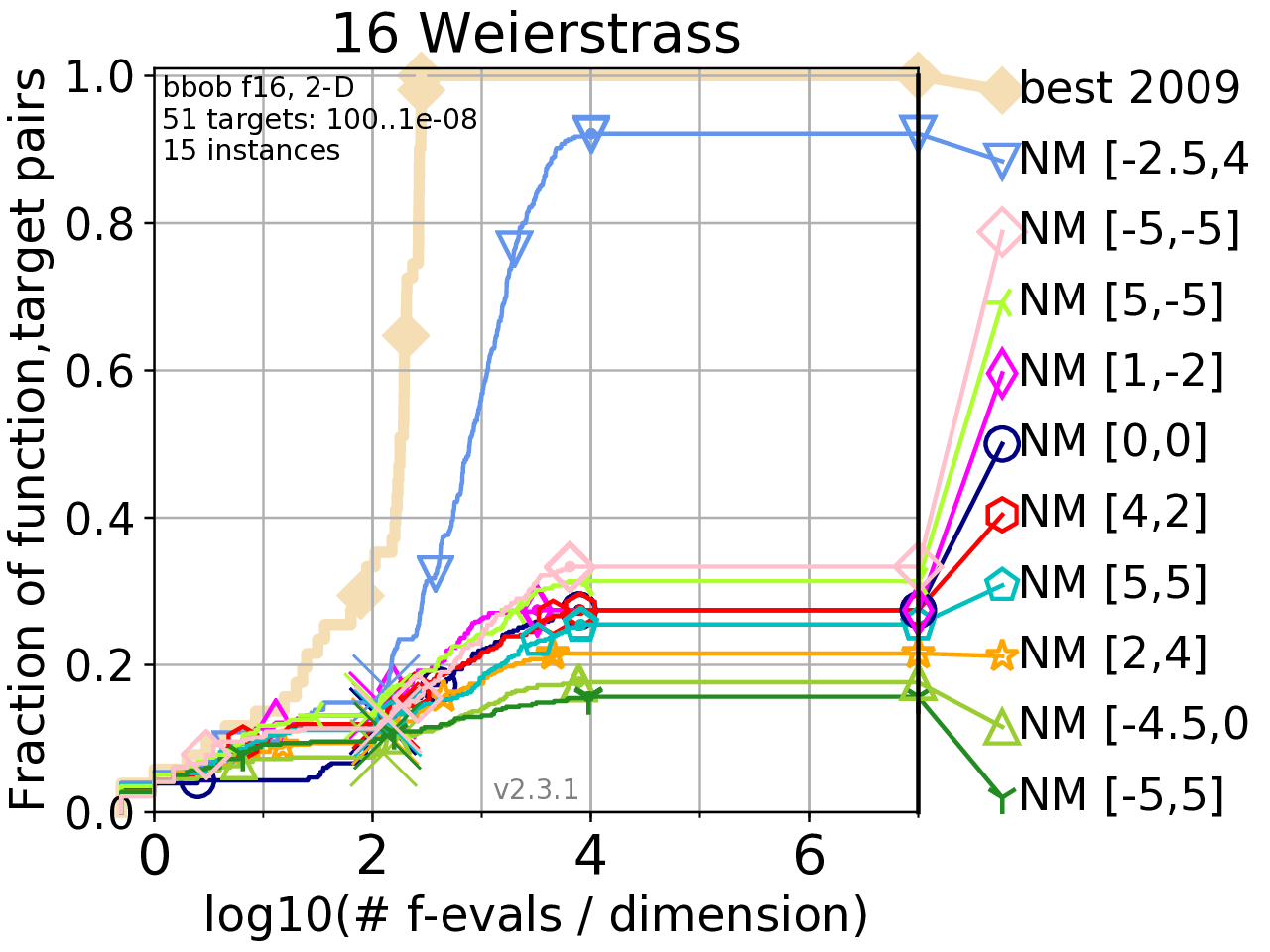}\hfill
    \includegraphics[width=0.43\textwidth]{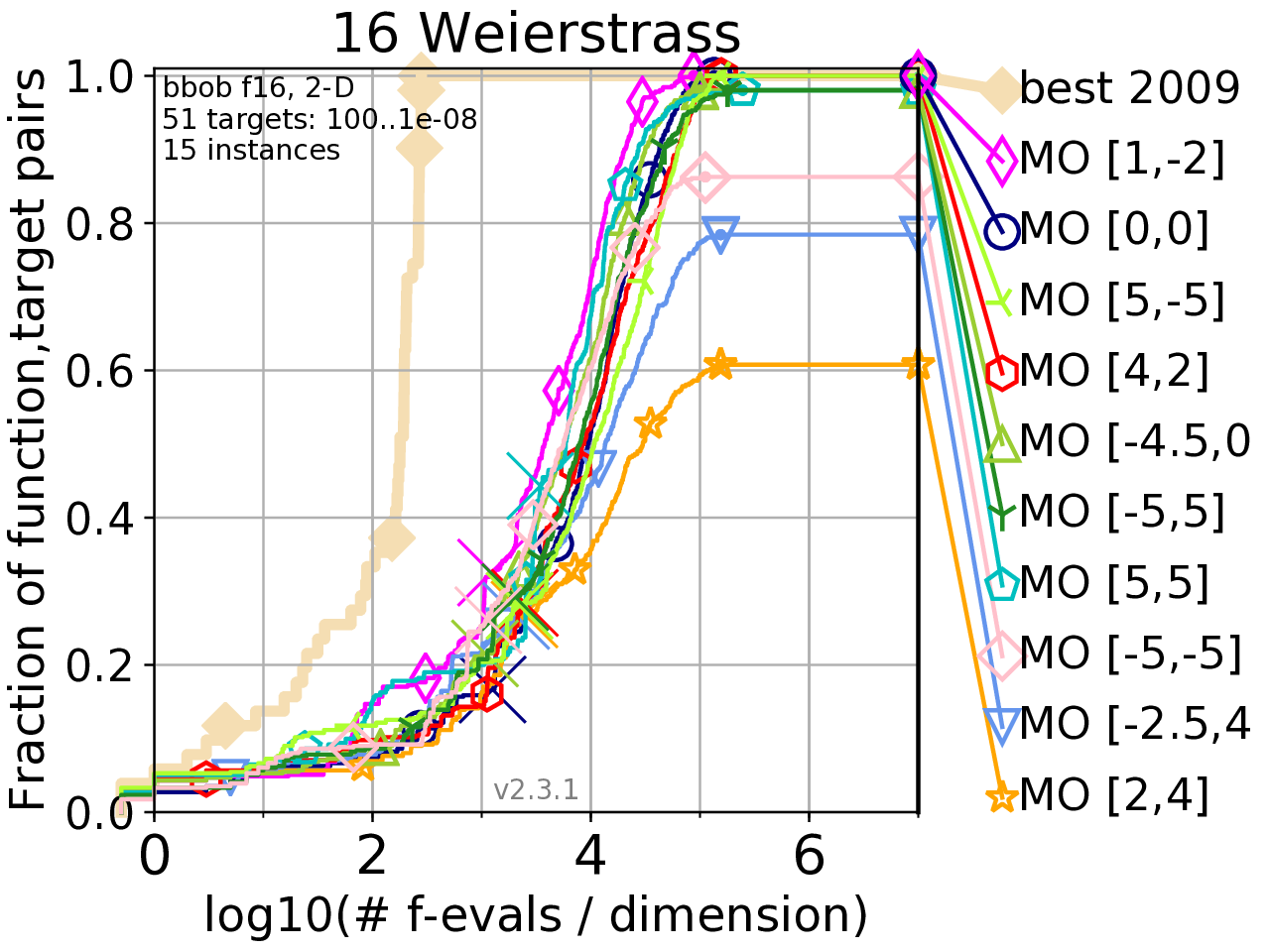}
    \caption{ECDF of runtimes on $f_3, f_4, f_{15},$ and $f_{16}$ in dimension 2 over 51 targets. On the left the runtime of Nelder-Mead from ten different starting points is displayed. The plots on the right show the same for MOGSA.}
    \label{fig:results_3-16}
\end{figure}

\begin{figure}[p]
    \centering
    \includegraphics[width=0.43\textwidth]{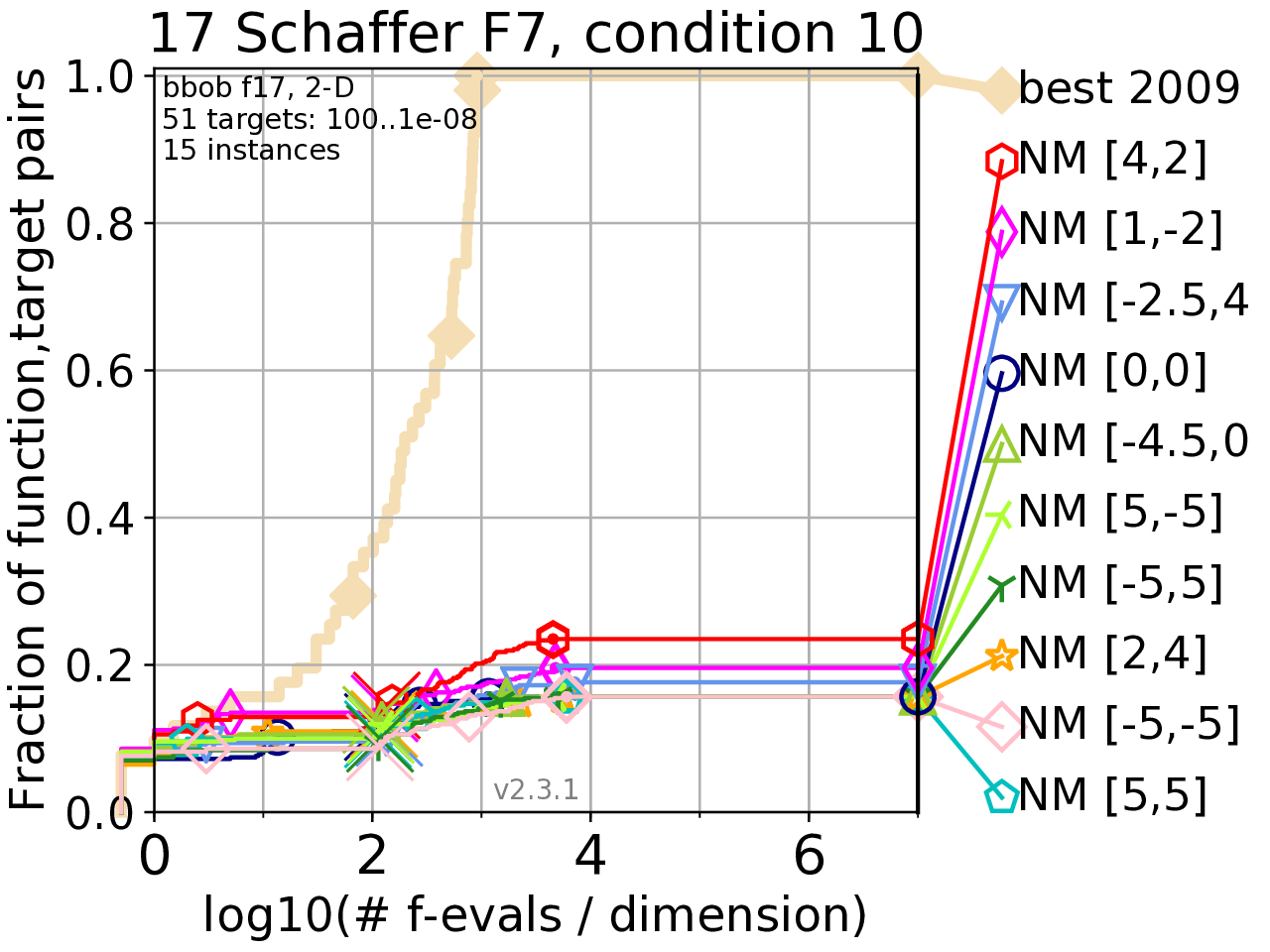}\hfill
    \includegraphics[width=0.43\textwidth]{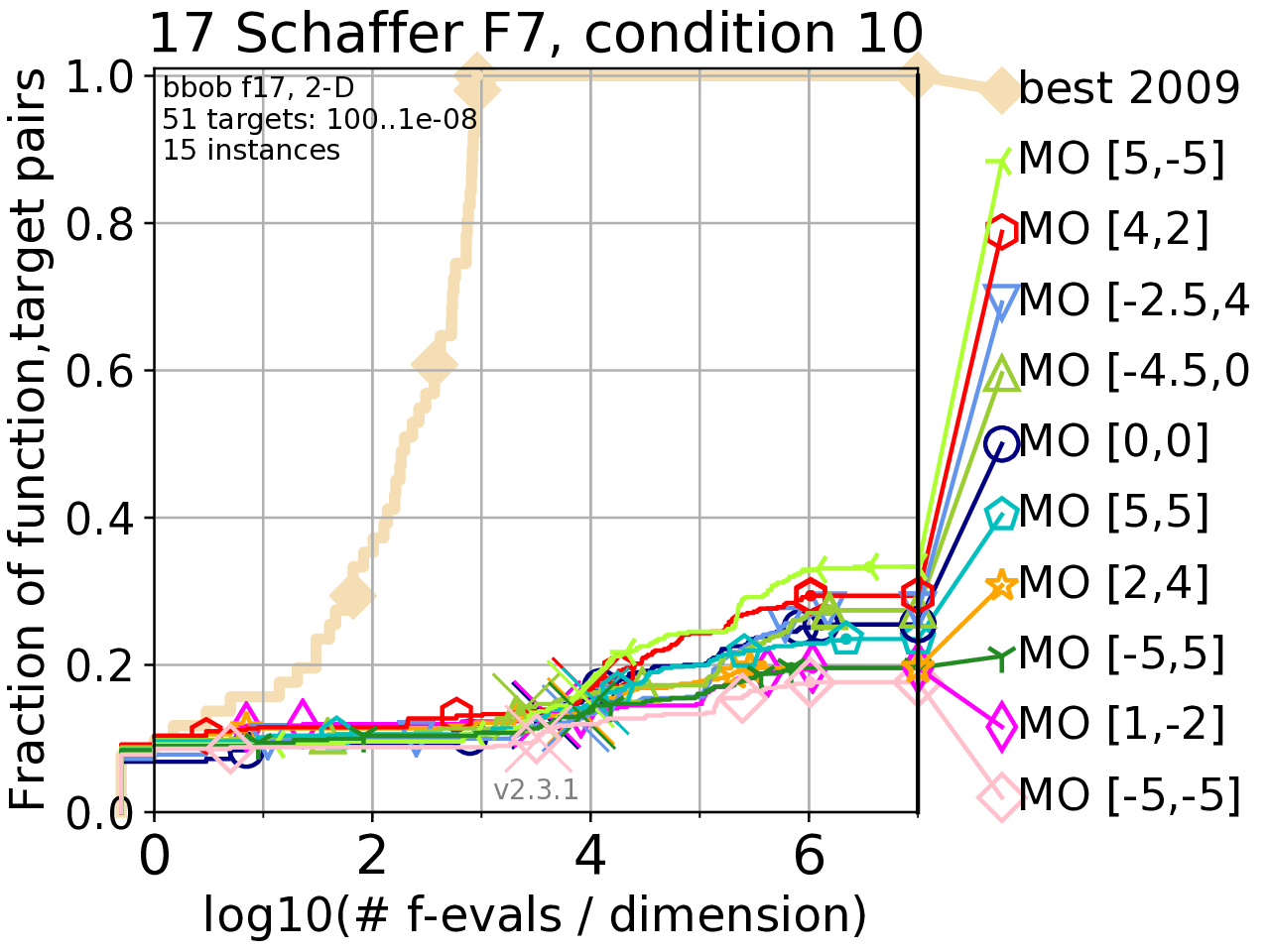}\\[0.25em]
    \includegraphics[width=0.43\textwidth]{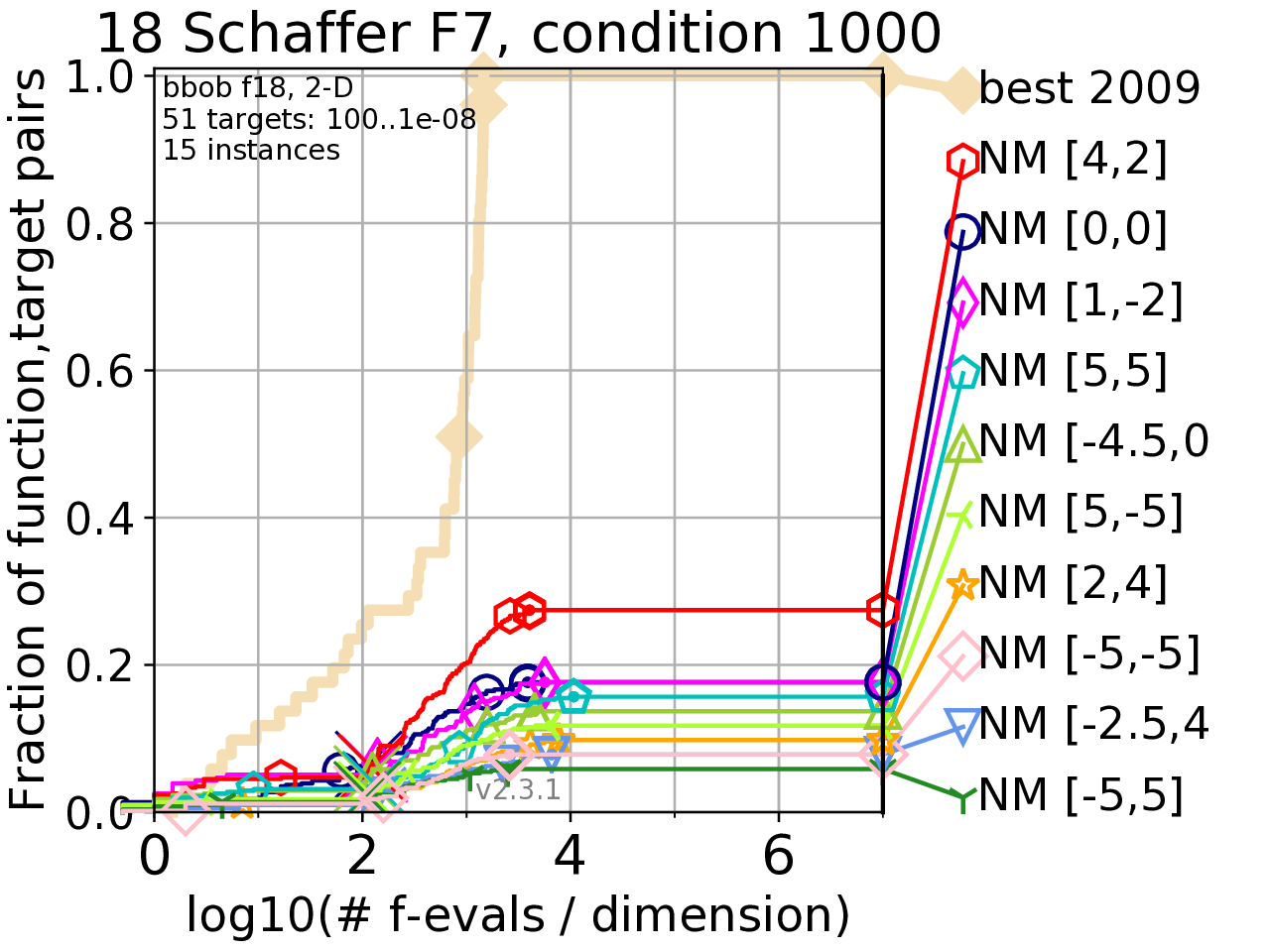}\hfill
    \includegraphics[width=0.43\textwidth]{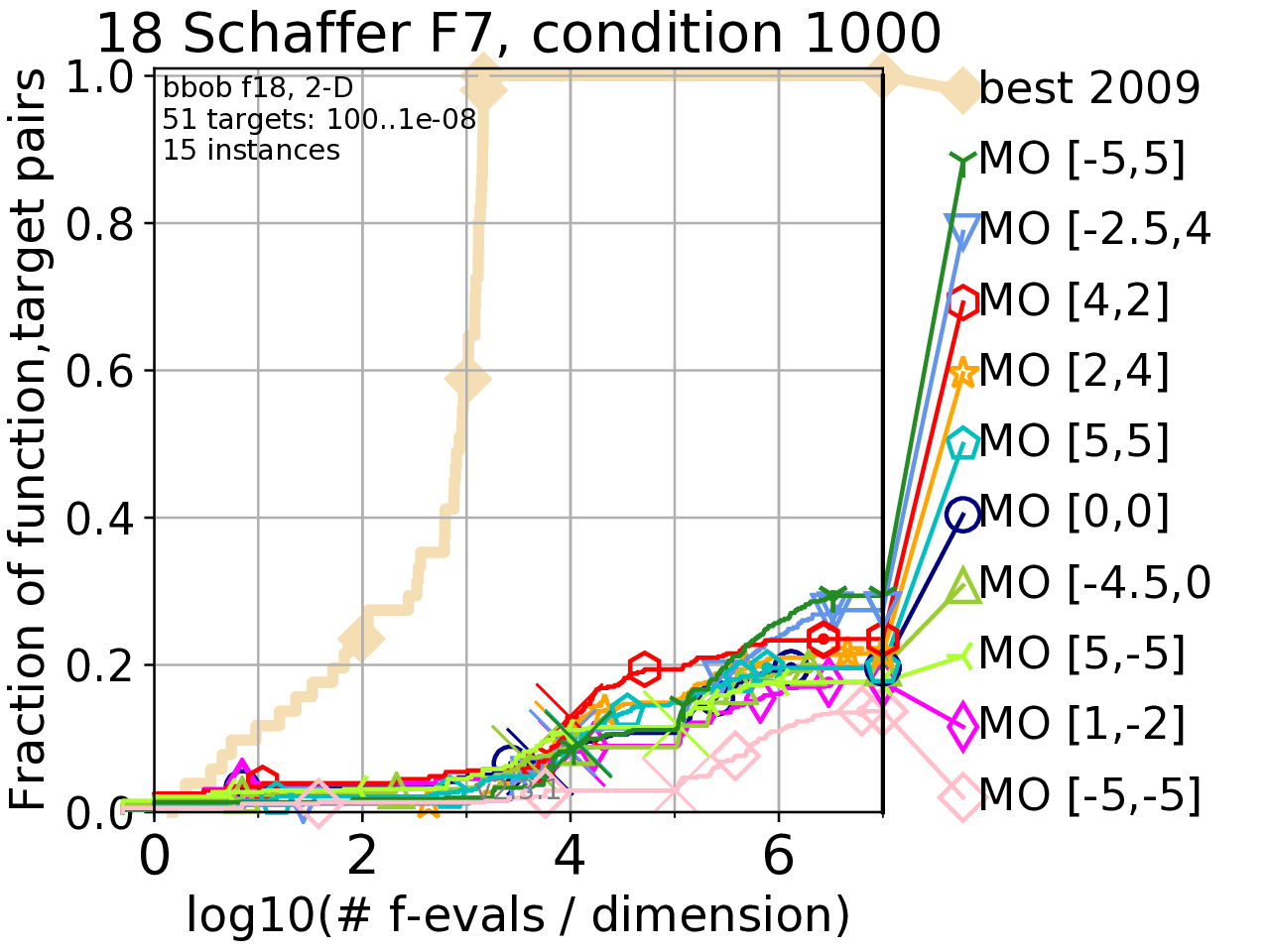}\\[0.25em]
    \includegraphics[width=0.43\textwidth]{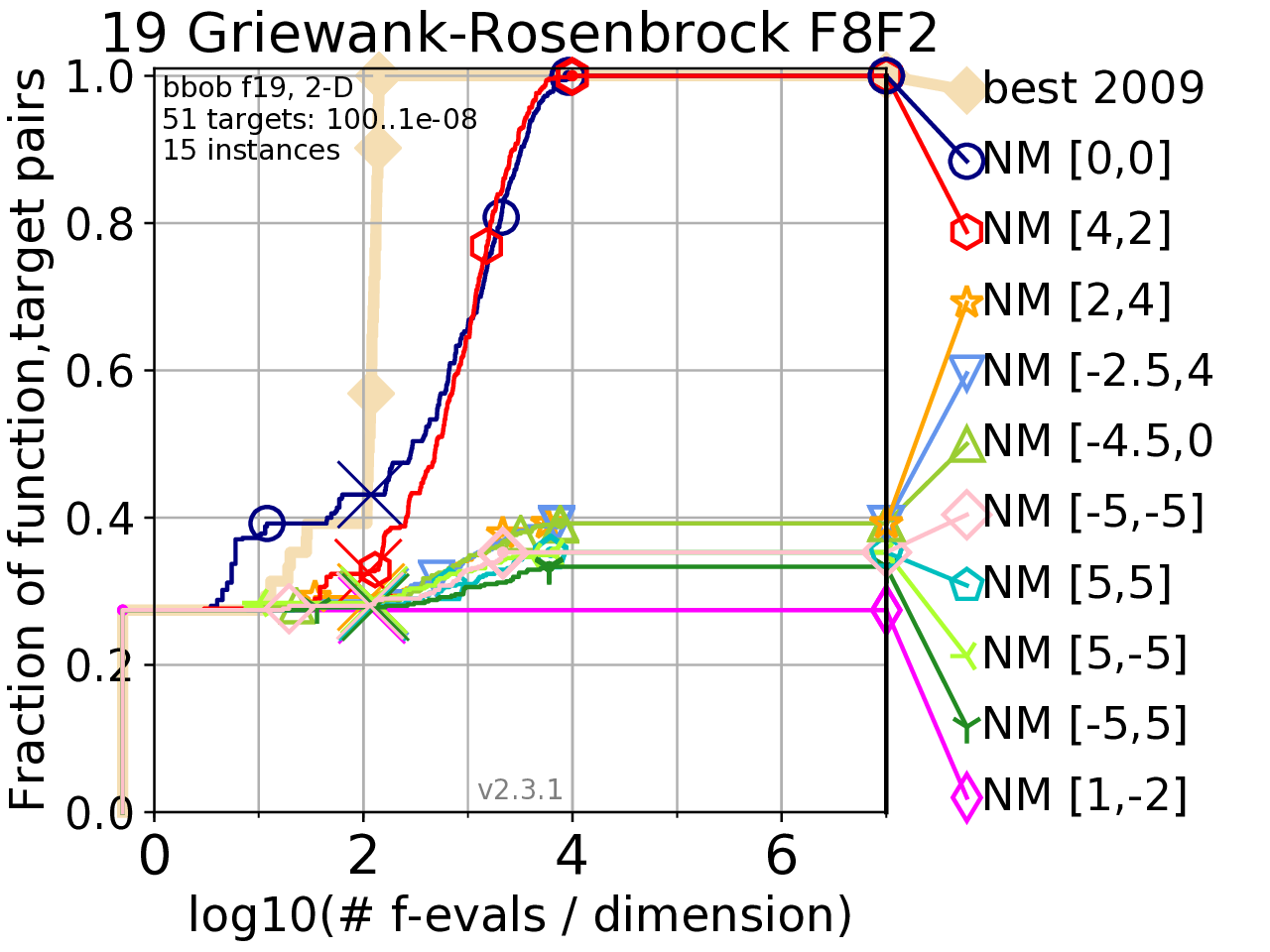}\hfill
    \includegraphics[width=0.43\textwidth]{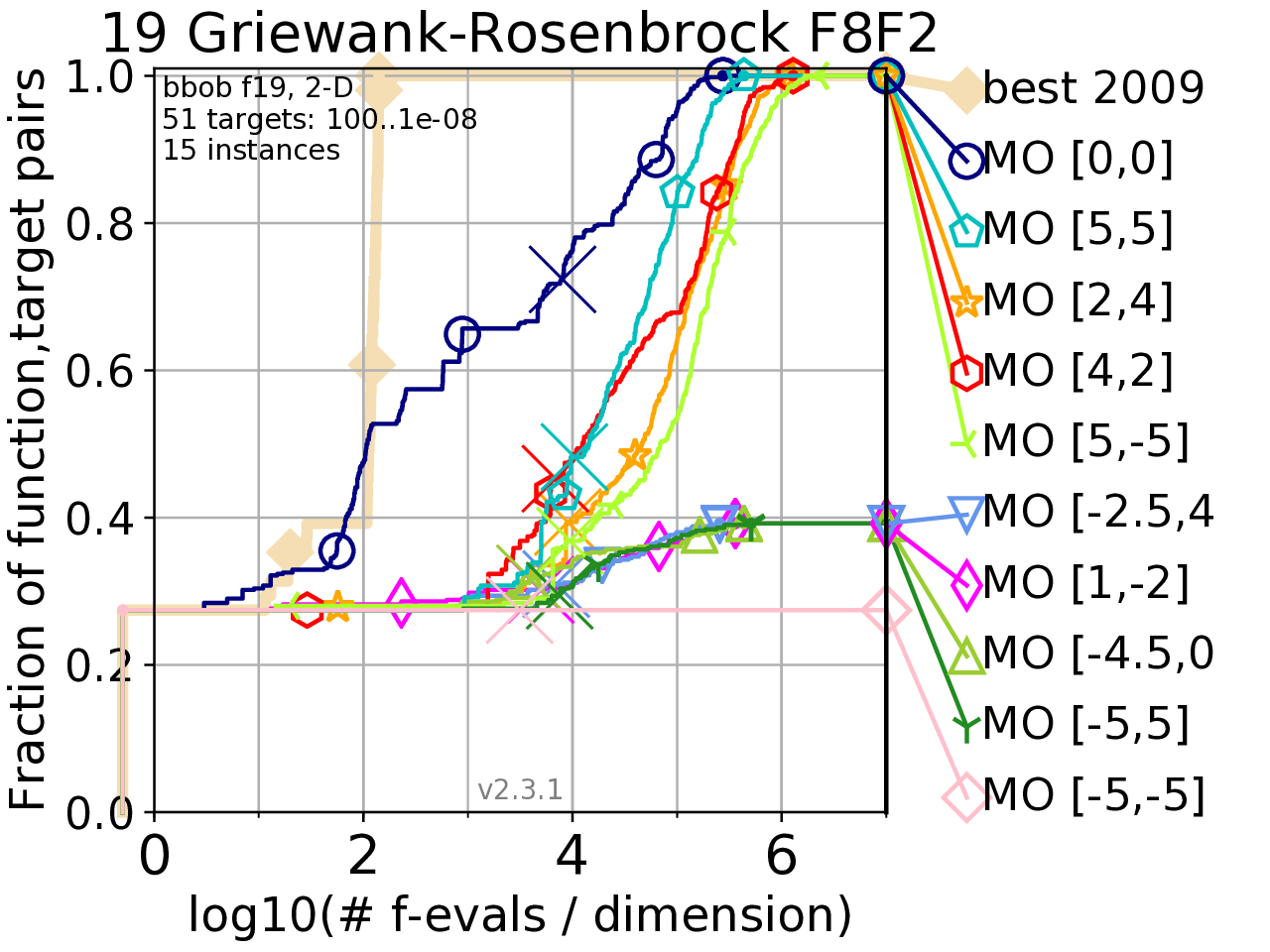}\\[0.25em]
    \includegraphics[width=0.43\textwidth]{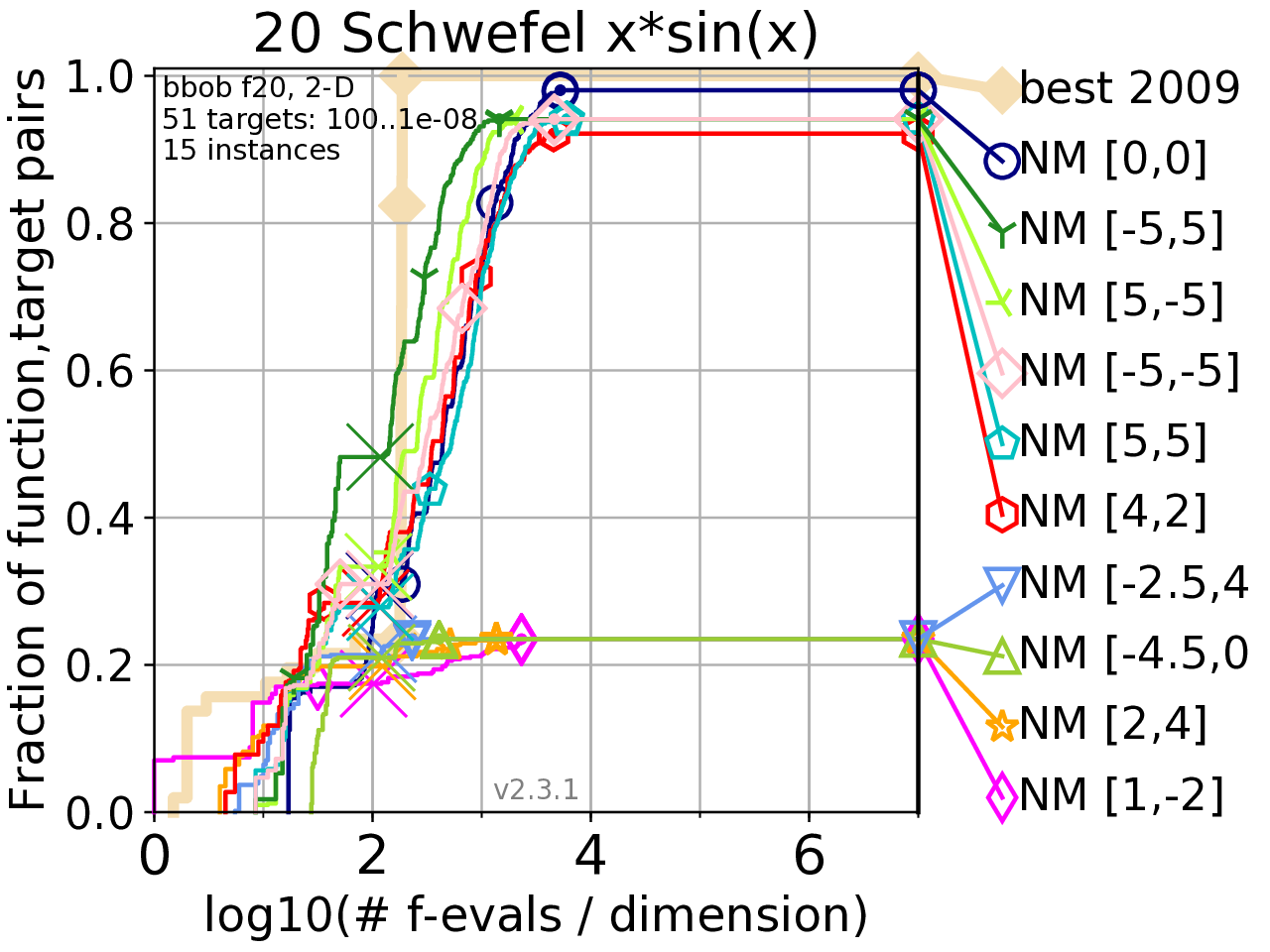}\hfill
    \includegraphics[width=0.43\textwidth]{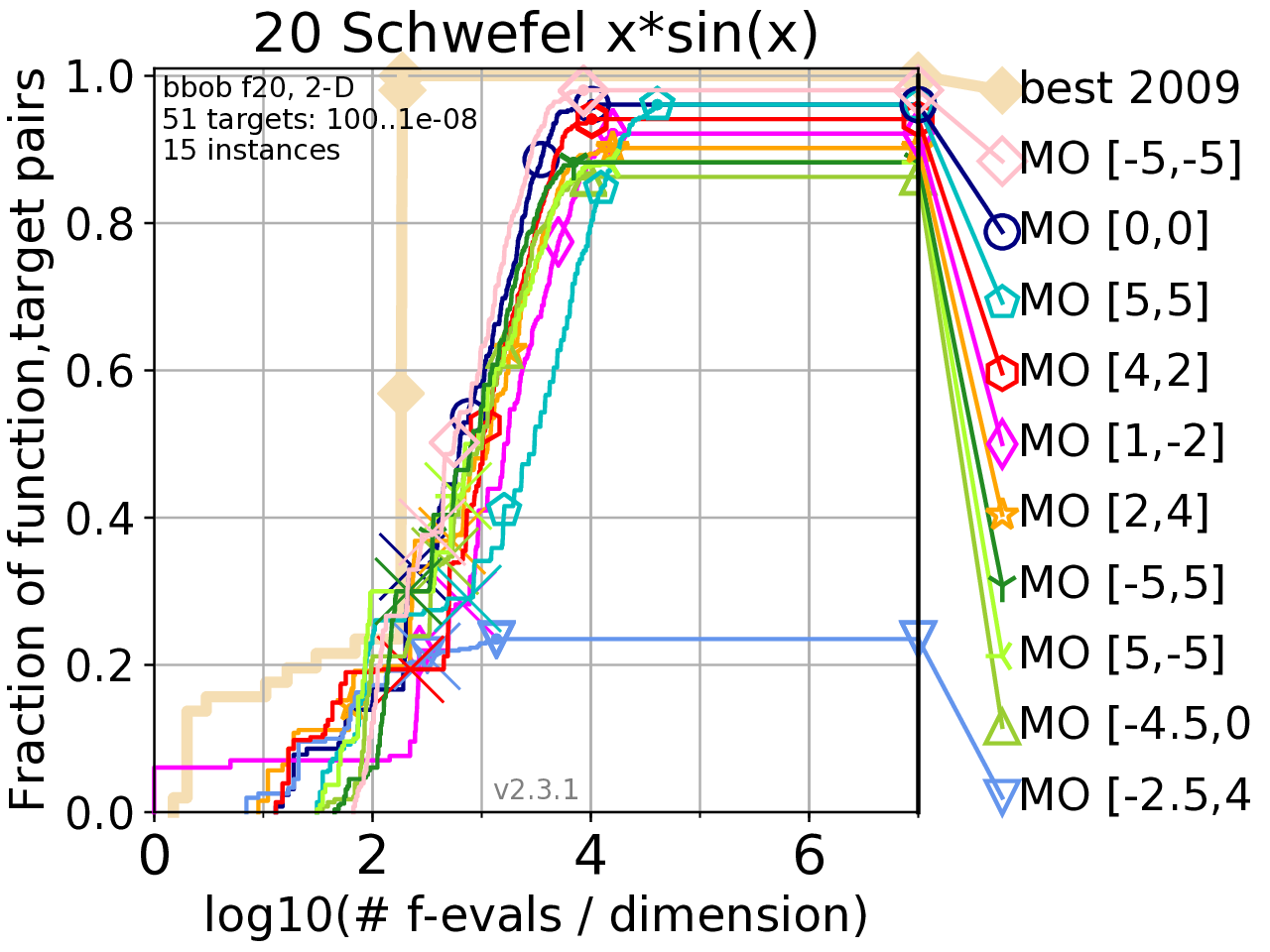}
    \caption{ECDF of runtimes on $f_{17}, f_{18}, f_{19},$ and $f_{20}$ in dimension 2 over 51 targets. On the left the runtime of Nelder-Mead from ten different starting points is displayed. The plots on the right show the same for MOGSA.}
    \label{fig:results_17-20}
\end{figure}

\begin{figure}[p]
    \centering
    \includegraphics[width=0.43\textwidth]{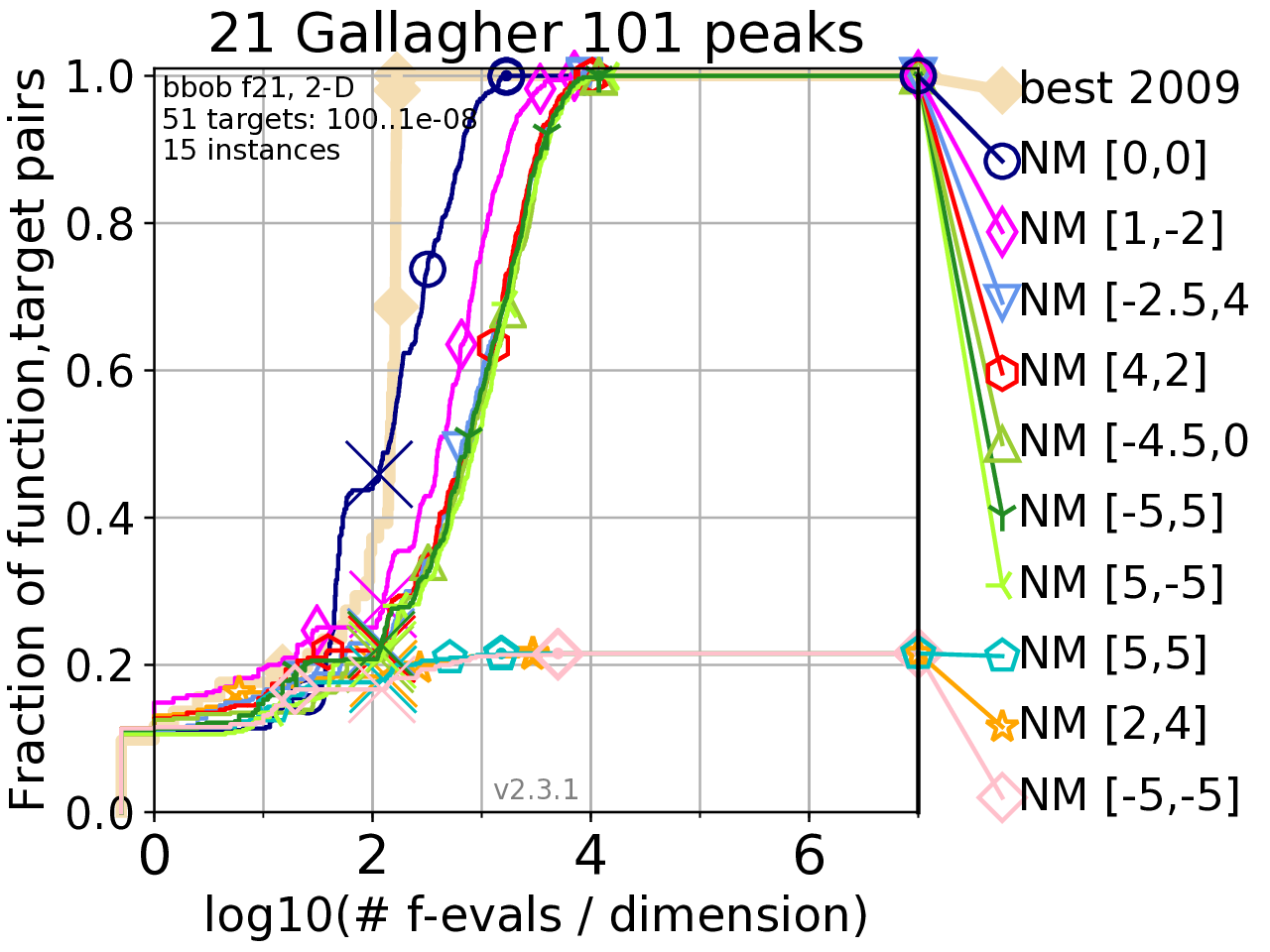}\hfill
    \includegraphics[width=0.43\textwidth]{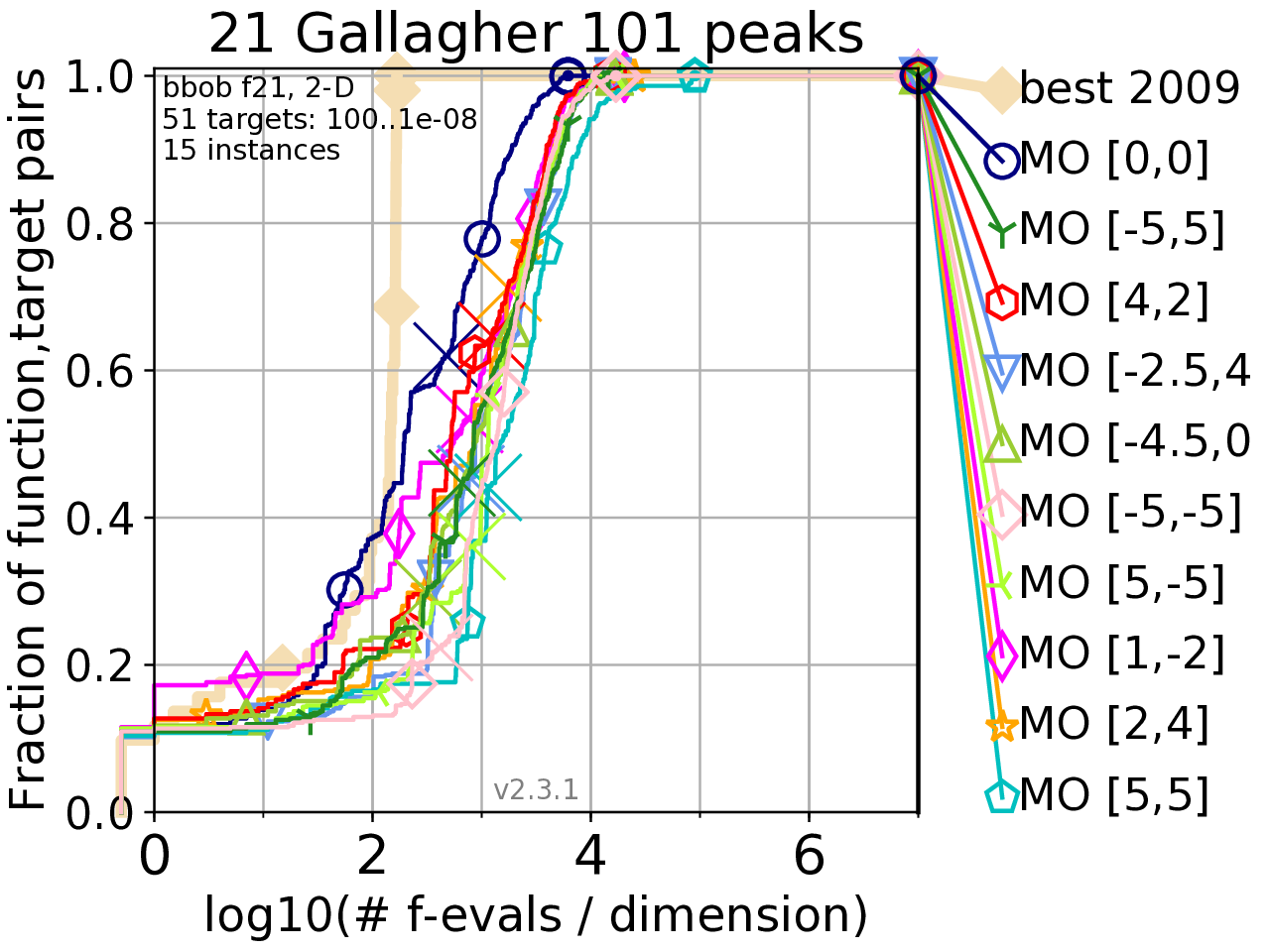}\\[0.25em]
    \includegraphics[width=0.43\textwidth]{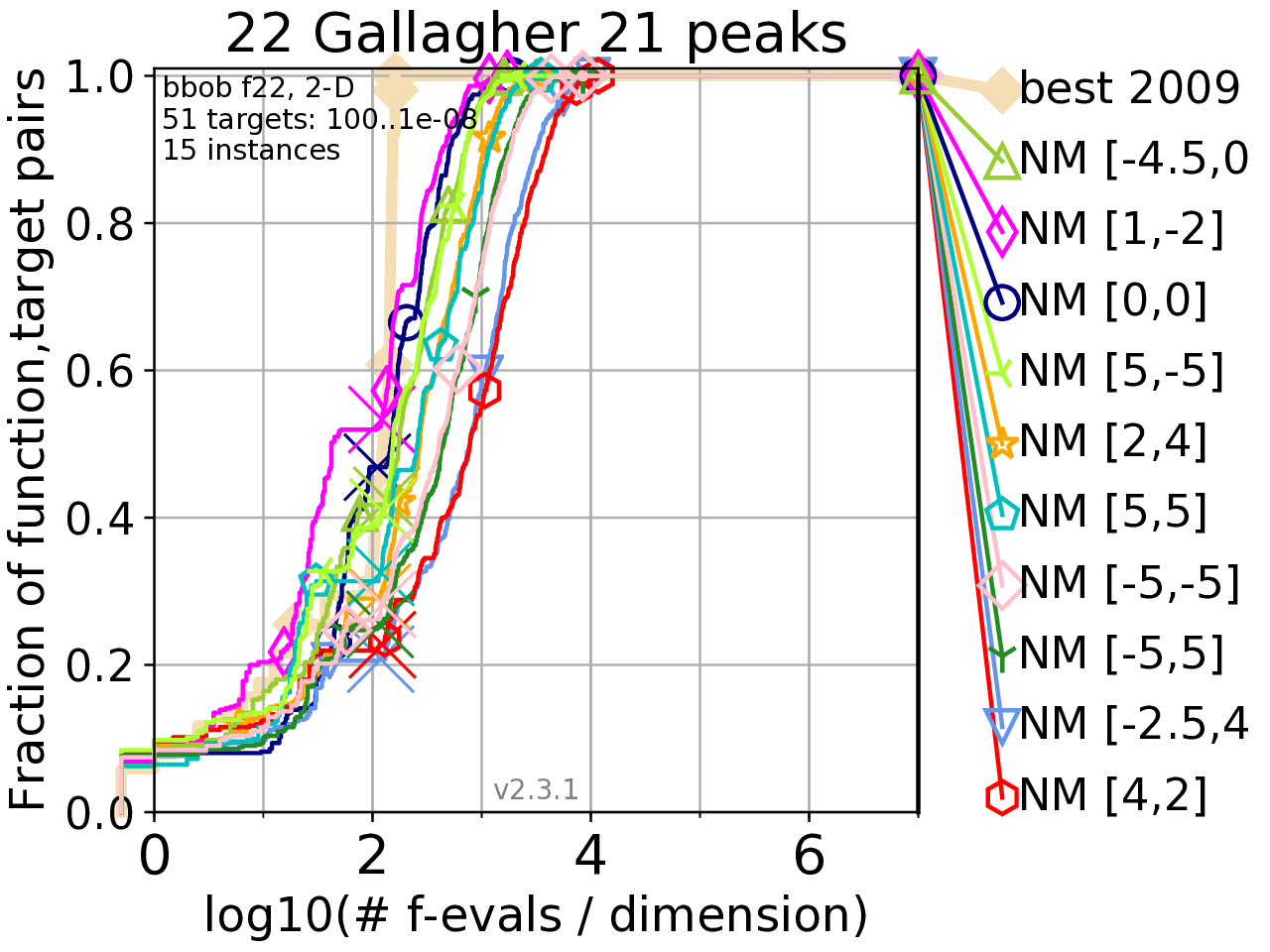}\hfill
    \includegraphics[width=0.43\textwidth]{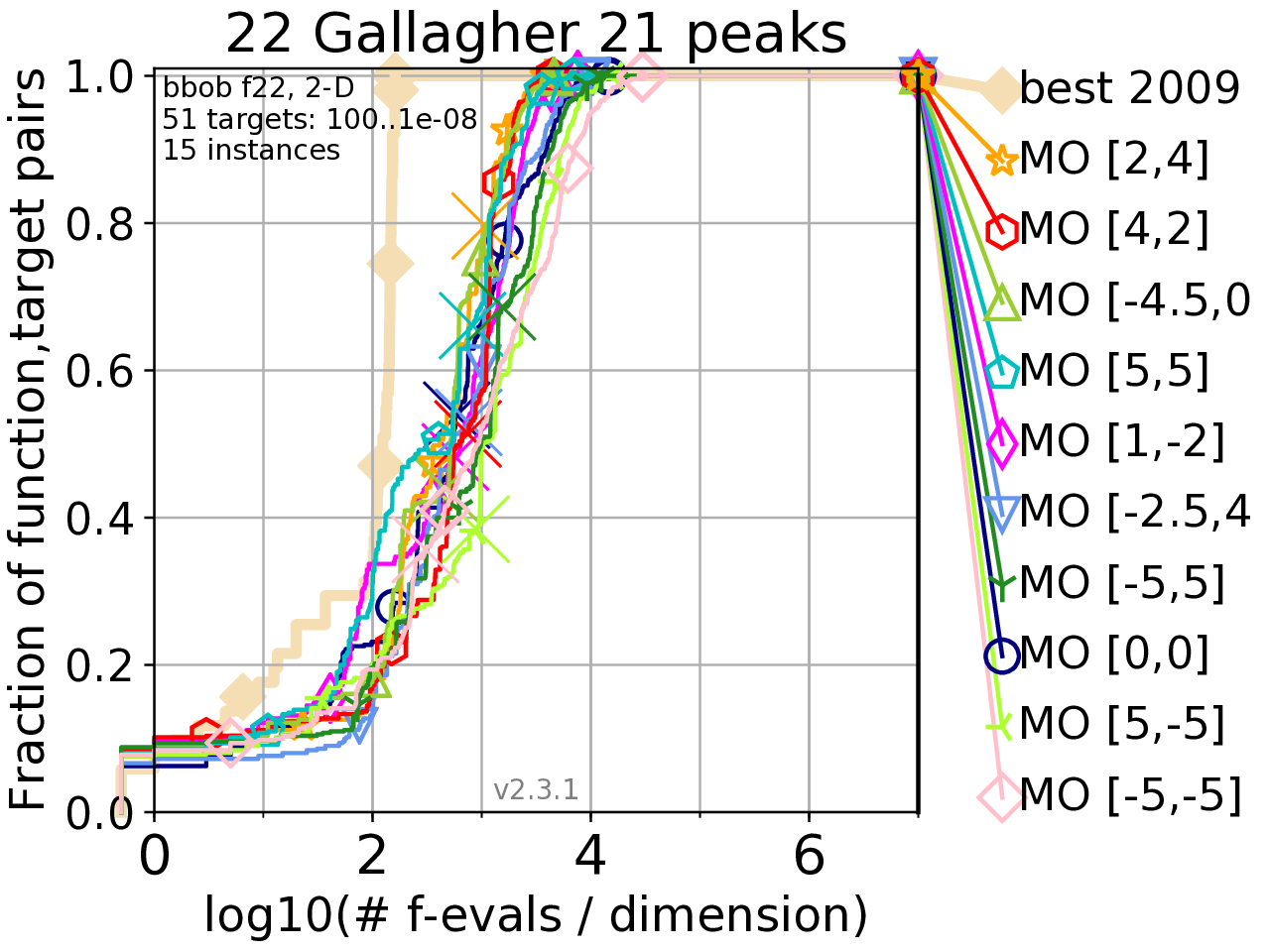}\\[0.25em]
    \includegraphics[width=0.43\textwidth]{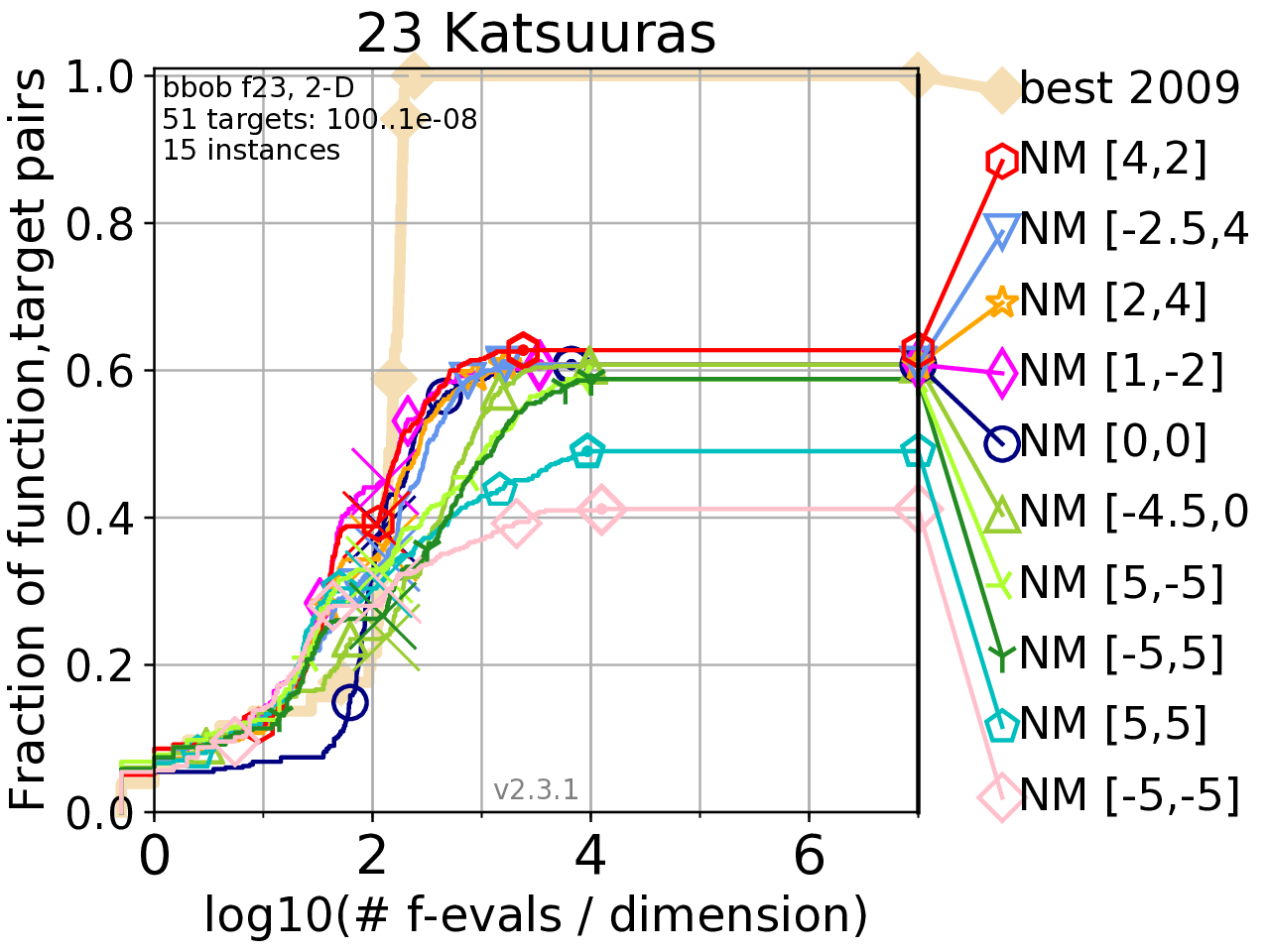}\hfill
    \includegraphics[width=0.43\textwidth]{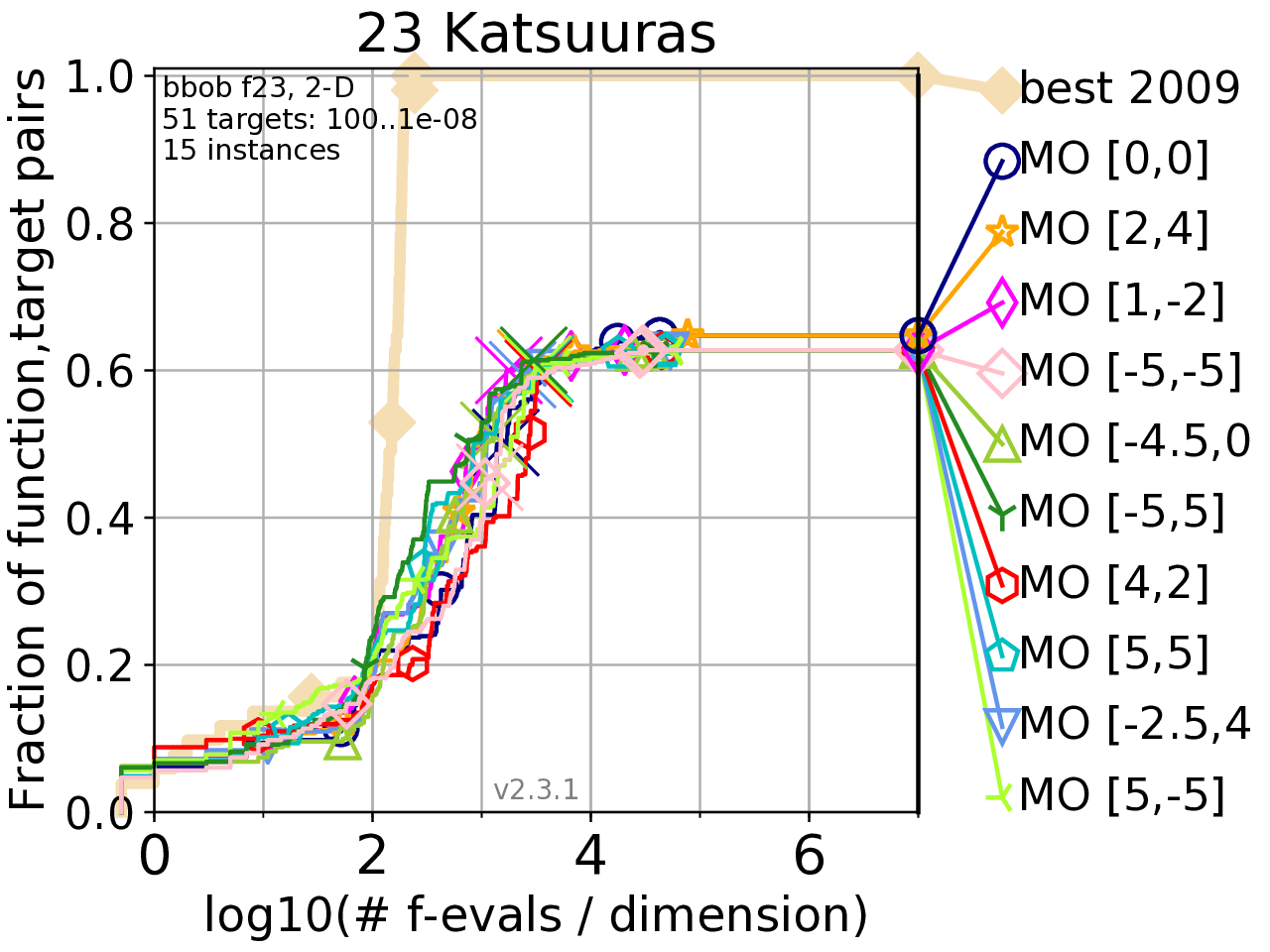}\\[0.25em]
    \includegraphics[width=0.43\textwidth]{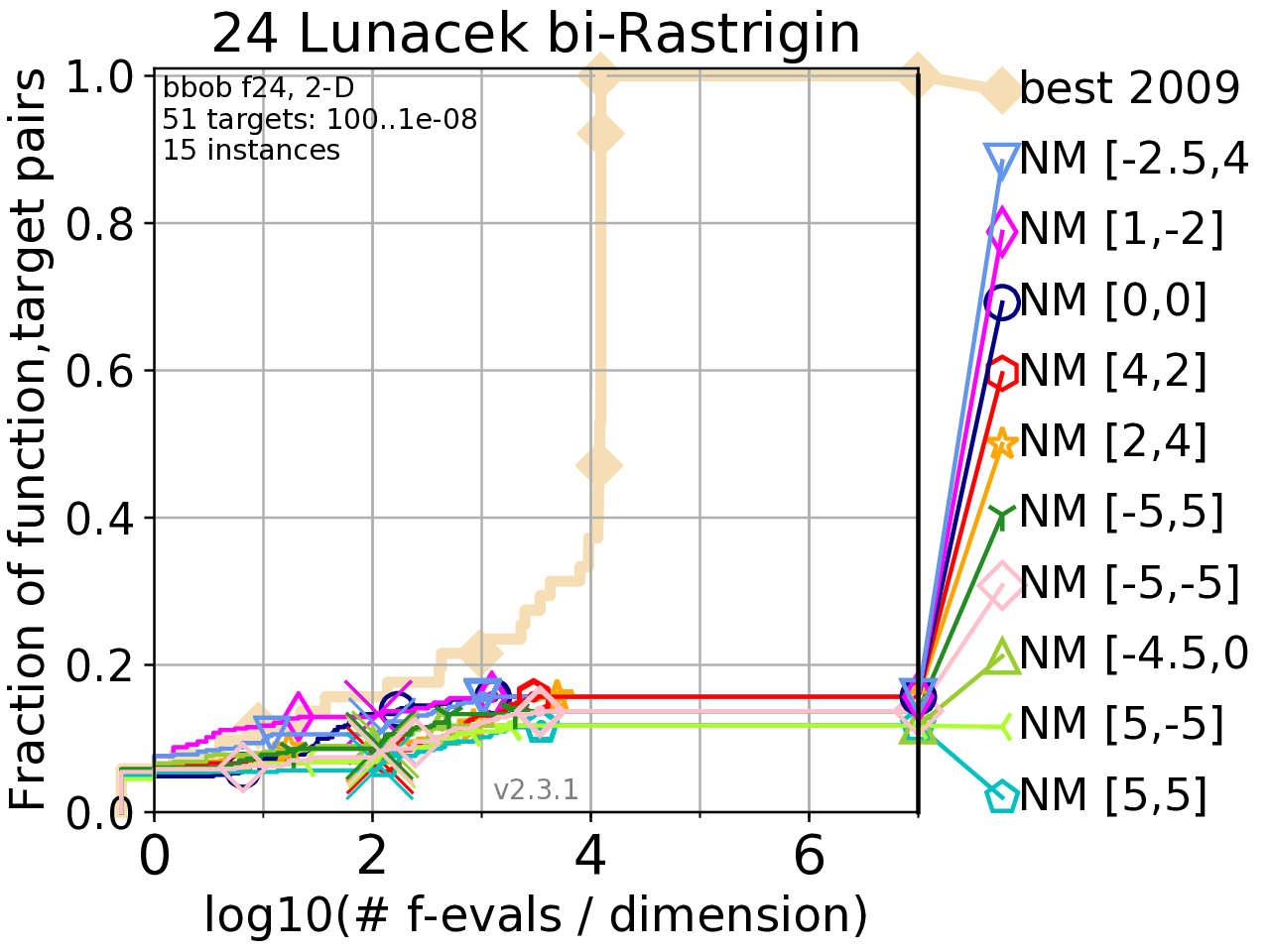}\hfill
    \includegraphics[width=0.43\textwidth]{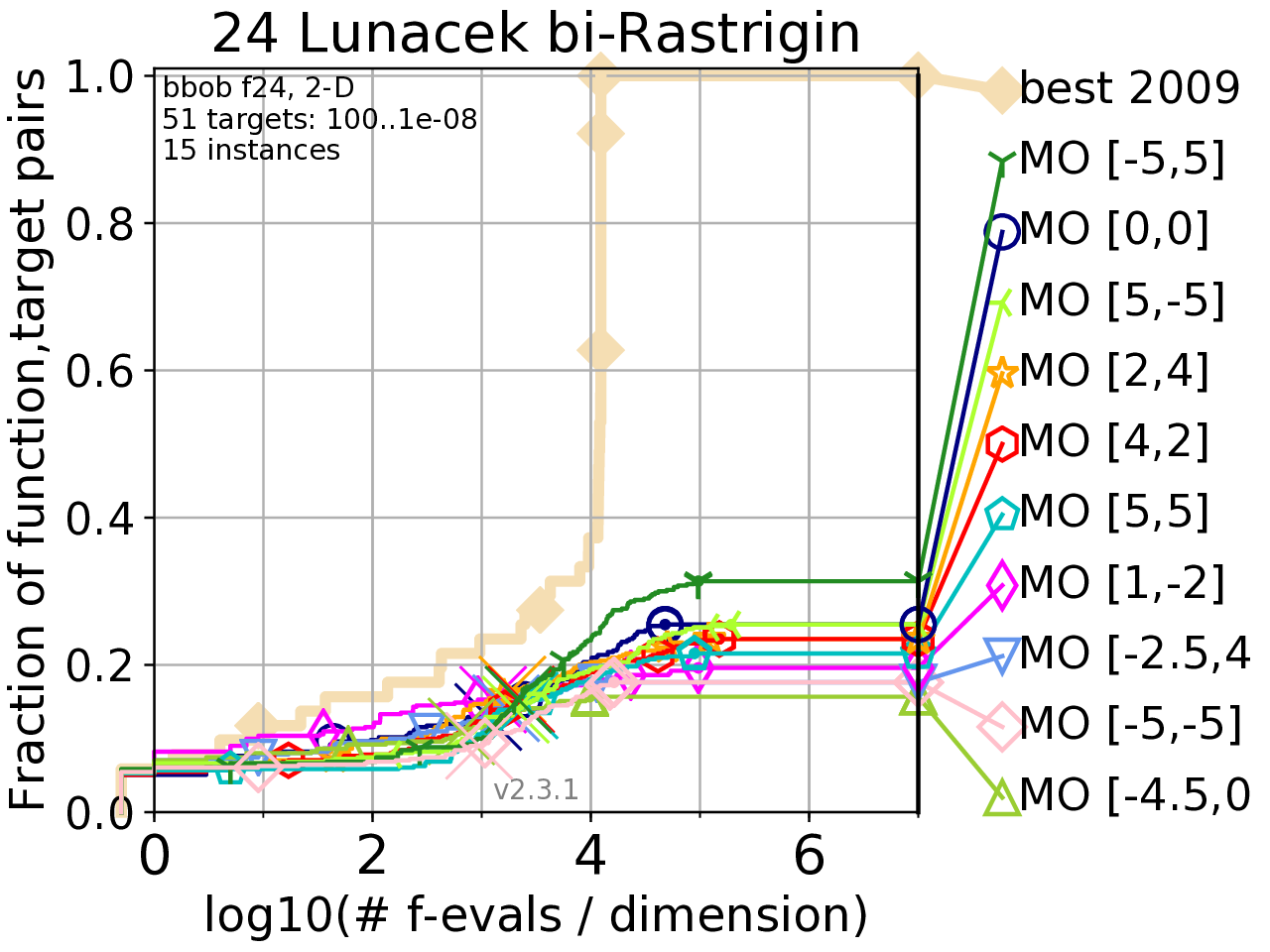}
    \caption{ECDF of runtimes on $f_{21}, f_{22}, f_{23},$ and $f_{24}$ in dimension 2 over 51 targets. On the left the runtime of Nelder-Mead from ten different starting points is displayed. The plots on the right show the same for MOGSA.}
    \label{fig:results_21-24}
\end{figure}

For analyzing whether the multi-objective landscape can be exploited for multi\-modal single-objective problems, we will focus on the multimodal problems of the COCO framework in the following. 
These include the two highly multimodal functions $f_3$ and $f_4$ from the separable function group, all multimodal functions with an adequate global structure $(f_{15} - f_{19})$ as well as all multimodal functions with a weak global structure $(f_{20} - f_{24})$. 
The respective plots of the conducted experiments in the COCO framework are presented in Figures~\ref{fig:results_3-16},~\ref{fig:results_17-20} and~\ref{fig:results_21-24}.

A success of applying MOGSA on the transformed single-objective problems is already evident in the upper three plots of Figure~\ref{fig:results_3-16}. %on which MOGSA was applied 
The separable Rastrigin function $f_3$, the Bueche-Rastrigin function $f_4$ and the non-separable Rastrigin function $f_{15}$ are structured and highly multimodal thus posing a lot of local traps.
Still, compared to Nelder-Mead (but also in total) MOGSA performed well. % concerning the results?
The plots for the runs of MOGSA on the right side reveal that the algorithm reached a value very close to the target function value with a precision of around $10^{-6}$ from six, for $f_{15}$ from seven of the ten starting values after $10^7\cdot 2$ function evaluations.
Nelder-Mead, on the other hand, reached that precision only for one, for $f_3$ for two starting points. All other runs solved only 10 to 20\% of the problems.
It is noticeable that MOGSA either solved around 80\% or only 10 to 20\% of the problems for different starting points. The latter was only the case for four, on $f_{15}$ for three of the runs.
This discrepancy between the reached precisions depending on the starting value is also observed for the highly multimodal Composite Griewank-Rosenbrock function $f_{19}$ in Figure~\ref{fig:results_17-20}. 
In every second run, 100\% of the problems were solved whereas in the other runs only 40\% were solved.
The same discrepancy can be identified for Nelder-Mead on this problem although with a lower number of successful runs.

Of all the considered multimodal functions, the Weierstrass function $f_{16}$, whose results are displayed in the bottom plot of Figure~\ref{fig:results_3-16}, is the one where MOGSA performed best compared to Nelder-Mead. 
The landscape of the Weierstrass function is highly rugged and moderately repetitive with a global optimum that is not unique.
For seven starting values, MOGSA solved the problems to 100\%. In the other three runs a precision of $10^{-4}, 10^{-6}$ and $10^{-7}$ was still reached which corresponds to having solved the problems to 60, 80 and 90\%.
The best run of Nelder-Mead starting in $(-2.5,4)$ solved 90\% the problems. In all the other runs, however, only 20 to 30\% of the problems could be solved.
The results indicate that Nelder-Mead eventually got stuck in local traps while MOGSA did not. While sliding from basin to basin, MOGSA can find the global optimum or at least optima with a function value very close to the best value.

On all the other multimodal functions MOGSA also reached better target precision values than Nelder-Mead. 
For $f_{17}$ and its counterpart $f_{18}$ at the top of Figure~\ref{fig:results_17-20} an overall higher percentage of solved problems can be observed after $10^7\cdot 2$ function evaluations.
Even on the Lunacek bi-Rastrigin function in Figure~\ref{fig:results_21-24} the values of the reached target precision, for the best run $10^{-1}$, were better for all runs of MOGSA than for those of Nelder-Mead.
On $f_{20}$, $f_{21}$ and $f_{23}$ the percentage of problems solved by Nelder-Mead differs between the runs.
MOGSA performs slightly better than the best run of Nelder-Mead but for all runs.
For $f_{22}$ both algorithms solved 100\% of the problems.

In summary, the plots of the ECDF of runtimes reveal that MOGSA is able to solve multimodal single-objective problems after having them transferred into multi-objective ones.
Noticeably, on some of the problems the result seemed to depend on the starting value.
However, even if the global optimum is not always found, an acceptable target precision is reached for almost all problems. 
After $10^7\cdot 2$ runs the best value found was always better than the one of Nelder-Mead for all multimodal problems.
Though MOGSA can definitely not be compared to the artificial best algorithm of BBOB-2009, the results are satisfactory especially considering that it is a local search and the problems provided possess a highly multimodal landscape as well as other difficulties.
To get an understanding of how these landscapes look like in a multi-objective setting, a visual analysis will be conducted in the following.

%examining the and the behavior of MOGSA on specific function instances 
%It can be concluded, that multiobjectivization can help to overcome local traps.
%- on some for different starting points completely different precision; depends on where MOGSA starts
%- take a look at specific function instances in the following and the way MOGSA takes

\subsubsection{Visual Analysis}
In the following, we will take a closer look at the heatmaps of individual function instances of the COCO framework after having added the sphere function $f_2$. Furthermore, we will visualize MOGSA's search trajectory based on its different starting points.

\begin{figure}
  \begin{subfigure}[b]{0.5\textwidth}
    \includegraphics[width=\textwidth, trim = 30mm 20mm 0mm 15mm, clip]{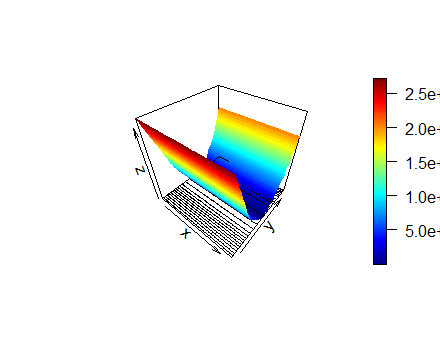}
    %\caption{single-objective problem }
  \end{subfigure}
  \hfill
  \begin{subfigure}[b]{0.4\textwidth}
    \includegraphics[width=\textwidth]{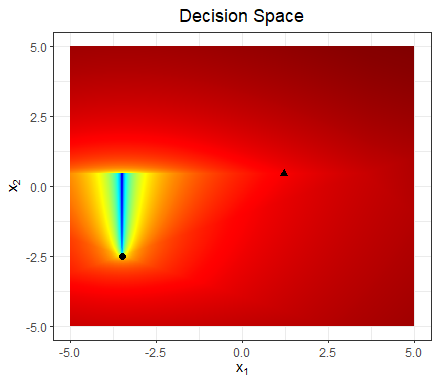}
   % \caption{multi-objective problem, joint visualization of an unimodal problem and a multimodal (two optima)}
  \end{subfigure}
  \caption{\label{fig:ellipsoidal} The left image displays an ellipsoidal function $f_1$ in a 3D plot. The heatmap on the right shows a bi-objective problem comprised of $f_1$ and an additional sphere function $f_2$. The global optima of both single-objective problems are depicted by the black triangle for $f_1$ and the dot for $f_2$.}
\end{figure}

As already stated above, our focus is not on unimodal functions. However, a visualization of an ellipsoidal function $f_1$ with an added sphere function $f_2$ (see Figure~\ref{fig:ellipsoidal}) reveals that the global optimum of the unimodal function $f_1$ is, other than expected, not located on the efficient set.
Figure~\ref{fig:ellipsoidal} depicts a 3D plot of the single-objective function $f_1$ on the left side. On the right side the decision space of the multi-objective problem is displayed in a gradient-based heatmap. 
As expected, the heatmap of the multi-objective problem being comprised of two unimodal functions shows only one basin of attraction and one efficient set. 
The objectives have no local optima that could constitute a trap and thus, no ridges exist. 
Nevertheless, only the optimum of the sphere function, indicated by the black dot, is located on the straight efficient set.
When looking at the 3D plot of $f_1$ one can see that the function has a broad valley.
There, the difference in height is rather small so that all gradients around the valley point into that valley instead of towards the best solution. 
Therefore, the efficient set in the multi-objective setting is always the path between the optimum of $f_2$ and the valley instead of the global optimum.
Thus, it is representing the shortest path from the optimum of $f_2$ towards the valley of $f_1$.
Since the behavior of MOGSA is based on the structure of the landscape, the algorithm walks as expected to that efficient set first. 
When reaching the end of the set by following $f_1$ and landing in the valley, the algorithm finds the global optimum of $f_1$ with the help of the local search.
The phenomenon of the optimum of $f_1$ not being located on the efficient set is observed for several unimodal functions, all with a rather broad valley.
Although our aim is not to solve unimodal problems, this shows that the global optima of both $f_1$ and $f_2$ are not always located on an efficient set, at least for unimodal functions.%\newpage

%could indicates that the re could be difficulties with multimodal functions as well. %%%%%%%%%%%%%%%%%%%%%%%%%%%%%%%%%%%%
%that the gradient-based landscape could look different than expected for multimodal problems as well. 
%/ this reveals that the efficient sets are located different than expected.

%of the coco framework ($f_2$, function instance 1)

\begin{figure}[t!]
	\centering
	\includegraphics[width=0.5\textwidth]{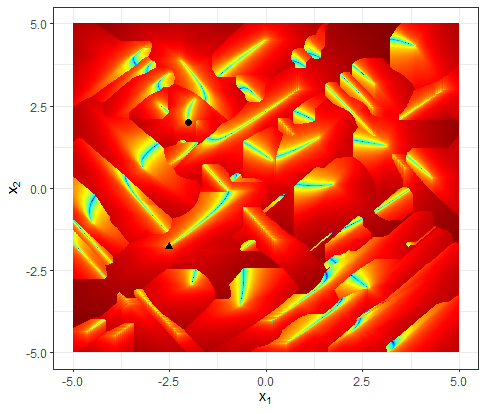}
	\caption{\label{fig_f21} Heatmap of the decision space of a multi-objective problem which is comprised of the Gallagher’s Gaussian 101-me Peaks function (in the COCO framework $p^3=(2, 21, 1)$) with its optimum illustrated by the black triangle, and the sphere function whose optimum is represented by the black dot.}
\end{figure}

For the multimodal problems of the COCO framework we see that the landscape after the multiobjectivization is sometimes also different than expected.
Figure~\ref{fig_f21} pictures the heatmap of the decision space of the Gallagher's Gaussian 101-me Peaks function $f_1$ with the sphere function $f_2$. 
Due to the multimodality a lot of basins of attraction as well as efficient sets that are cut by a ridge are visible. 
Only one efficient set is ridge-free.
Surprisingly, only the global optimum of $f_2$ is located on that set, illustrated by the black dot.
Wherever MOGSA starts, it ends in that set by exploring the other efficient sets, jumping from basin to basin when facing a ridge until the ridge-free set is found.
However, the global optimum of the first function (black triangle) is located on the end of another set that is cut by a ridge.
Whether MOGSA finds that global optimum or not depends on where it starts. 
When the respective efficient set is on the way that MOGSA takes, it will explore the set, find and save the global optimum of $f_1$.
Although the algorithm does not regard the point found as globally efficient at first due to the existing ridge, it recognizes the globality when comparing all saved points before stopping. 
Still, it is possible that MOGSA does not run in the basin of that set and thus does not find the global optimum of $f_1$.
This would be the case when the starting point is for example in the upper part of the heatmap.
When starting somewhere in the lower part, for example in $(-1, -3)$, it finds the global optimum of $f_1$ on the way to the ridge-free efficient set.
Based on this heatmap we can state that the global optima of a bi-objective problem can be located on different efficient sets and thus in different basins of attraction.
As already assumed during the evaluation of the results generated by the COCO framework, the heatmap also indicates that the starting point has an influence on the success of the optimization process.

%On multimodal problems, different than expected global efficient set not always visible, global optima of both functions are located on different efficient sets (in different basins of attraction):
%global efficient set on which the global optima of both functions are located does not exist

\begin{figure}[t!]
	\centering
	\includegraphics[width=0.75\textwidth]{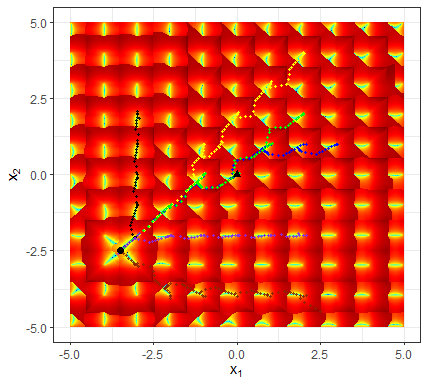}
	\caption{\label{fig:rastr4} Heatmap of the decision space of a multi-objective problem which is comprised of the Rastrigin function $f_1$ with its optimum illustrated by the black triangle, and the sphere function $f_2$ whose optimum is represented by the black dot. The colored dots on the heatmap show the way of MOGSA starting in six different points.}
\end{figure}

Figure~\ref{fig:rastr4} confirms this theory. The heatmap depicts a multi-objective problem which is comprised of the Rastrigin function $f_1$ and the sphere function $f_2$. The global optima are represented by the black triangle for $f_1$ and the dot for $f_2$.
Since $f_1$ has a lot of local optima, many efficient sets exist that are all oriented towards the global optimum of $f_2$.
As in the previous heatmap, a ridge-free global efficient set on which the global optima of both single-objective functions are located does not exist.
Contrary to the optimum of $f_2$, the global optimum of $f_1$ is not visible and no difference to the other efficient sets can be seen.
The behavior of MOGSA on this multi-objective problem is depicted by the colored dots.
The blue land represents the way for starting in $(3,1)$, yellow for $(2,4)$, green for $(2,2)$, black for $(-3,2)$, dark red for $(2.5,4.5)$, and purple for $(2,-2)$.

On the given landscape, MOGSA behaves as expected for all starting points. 
First, the efficient set of the respective basin of attraction is found which will be explored in the second phase. 
By following the gradient of $f_1$, an optimum is found and the gradient of $f_2$ is followed. This is done until a ridge cuts the set or the global optimum of $f_2$ is found. 
For some ways we detect that the path MOGSA takes deviates slightly from the efficient sets, for example for the blue and the dark red path.
However, if the set is only left a little it does not have an impact on the search as MOGSA still follows the single-objective gradient, in the cases of the deviation the one of $f_2$. 
This means that the algorithm is still walking towards the optimum to which the efficient sets are not pointing as direct as the algorithm walks.
The next ridge is found regardless of small deviations from the efficient set.
In case of no ridge but an optimum, there are no deviations and MOGSA walks straight to that optimum.

Regardless of the starting point, the global optimum of $f_2$ is found in each of the runs as all efficient sets point towards that optimum where the search stops.
The global optimum of $f_1$ on the other hand, is only found when the respective efficient set is located on the way towards the optimum of $f_2$. 
From the six starting points this is only the case for the green and the blue path. 
MOGSA is not able to walk up the ridges so that once it left a basin it is not able to go back. 
Thus, the success of the search highly depends on the starting point.

\begin{figure}[t]
    \centering
    \includegraphics[width=0.44\textwidth]{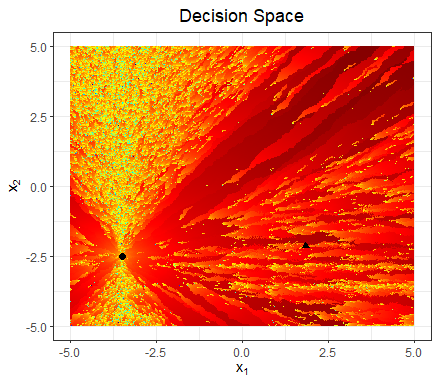}\hfill
    \includegraphics[width=0.44\textwidth]{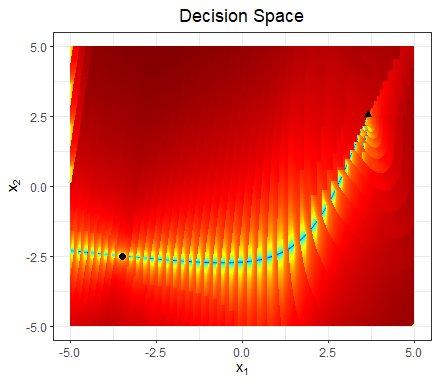}\\[0.25em]
    \includegraphics[width=0.44\textwidth]{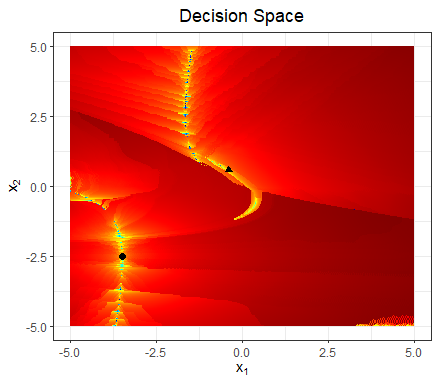}\hfill
    \includegraphics[width=0.44\textwidth]{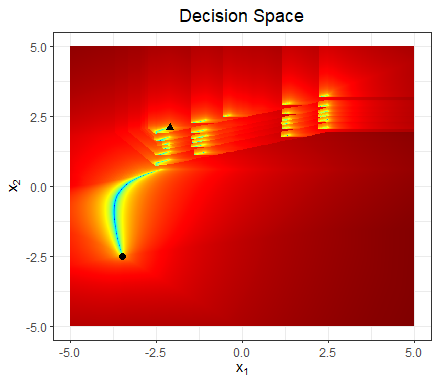}
    \caption{Heatmaps for the decision space of different problem instances of the COCO framework with an added sphere function $f_2$ whose optimum is located in the black dot. 
    The problem instances: $p^3=(2, 16, 1)$ at the top left next to $p^3=(2, 17, 1)$, at the bottom left $p^3=(2, 19, 3)$ and on the bottom right $p^3=(2, 20, 1)$. Their optima are illustrated by the black triangle.}
    \label{fig:heatmaps_coco}
\end{figure}

%\newpage
Both discussed problems are highly multimodal each with further characteristics. 
Considering the other multimodal problems from the COCO framework we can say that all of them have serious difficulties that MOGSA has to face.
Some of these problems are depicted in the heatmaps of Figure~\ref{fig:heatmaps_coco} after the multiobjectivization.
They underline our observation that the two global optima of bi-objective problems are not always located on the same efficient set. 
Sometimes one or even both global optima are not even visible like in the heatmap at the bottom left.

%On more complex multimodal problems, even sophisticated algorithms struggle to find the global optimum due to the local ones.
%No derivatives – approximation of the gradients
%\newpage

\section{Conclusion}\label{sec:concl}
Multimodality is a crucial factor defining the hardness of single-objective problems. 
With our approach of transforming single-objective into multi-objective problems to exploit the properties of their landscape, we show a way to deal with local optima. 
In the multi-objective setting, local optima do not pose a threat anymore as the optimizer MOGSA is able to overcome the traps and even benefit from local optima, which guide the algorithm to global efficient sets. %on its way to the global efficient set.
Although the functions on which we conducted our experiments possess severe difficulties as shown in the visualizations, the algorithm was able to find global optima or reach a function value close to the final target value of $f_1$.
On some problems, however, the structure was too complicated. 
Nevertheless, one should not forget that MOGSA is a deterministic local optimizer and even if the global optimum of $f_1$ was not found, our experiments reveal that it is not because of local traps.
With the ability to slide through the landscape of multi-objective problems exploring efficient sets, MOGSA is visiting several optima on its way without getting stuck. 
In the gradient-based heatmaps we observe that global optima are not always visible, at least not for both single-objective functions.
In this case, it often depends on the starting position of the algorithm which always finds the way to the optimum of $f_2$.
If the global optimum of $f_1$, the single-objective problem we want to optimize, is found or not depends on whether it is located on the path MOGSA takes. 

This opens up opportunities for future work. 
With the knowledge of being able to find the global optimum from specific starting points, the choice of the \emph{right} starting point could be analyzed and improved. 
Possibly, the application of an evolutionary algorithm could help to select the best starting point after using the mechanisms of mutation and recombination.
%The point with the best function value on the path towards the optimum of $f_2$ could be saved and used for these mechanisms to find a point from which the global optimum is located on the way of MOGSA.
Another way of building up on this work is an analysis of the location of the additional second objective $f_2$. 
Having shown that multiobjectivization can also be beneficial for other than evolutionary algorithms, a dynamic environment could improve the local search as well. 
The concept mostly applied in evolutionary optimization for maintaining diversity~\cite{branke2012dynamic} could be transferred to our local search. 
Still aiming at optimizing a constant function $f_1$, the location of the additional objective $f_2$ could be changed during the search process. 
Without getting stuck in local traps, the diversity could lead to the exploration of a broader area of the multi-objective landscape.
%many optima could be visited by exploring the efficient sets and better function values achieved, in the best case the one of the global optimum of $f_1$.
%Thus, a bigger area of the landscape could be explored and better function values achieved, in the best case the one of the global optimum of $f_1$.
%- analyzing the changing landscape depending on the additional objective
% By maintaining a sample of previously visited areas in the search space while improving still the quality of evolved solutions, the time for adaptation decreases when the environment changes. A variety of methods have been proposed in the literature by maintaining diversity, maintaining a separate memory to store the best solutions found in each generation, or by using multi-populations to track areas in a changing landscape.
Further research can be conducted in the field of optimizing MOGSA and its parameter-setting. Especially the step-size could be focused on for ideally adjusting the size automatically to the landscape.
%- Adapt step size of MOGSA, development of sophisticated step-size adaption mechanisms, which ideally adjust the parameters automatically to the landscape at hand

All in all, we have shown that -- by adding an additional objective -- multiobjectivization can be beneficial for multimodal continuous single-objective optimization and highlighted potential for further investigation. 

\subsection*{Acknowledgments}
The authors acknowledge support by the \href{https://www.ercis.org}{\textit{European Research Center for Information Systems (ERCIS)}}.

\bibliography{somogsa}
\bibliographystyle{abbrv}
\end{document}